\pdfoutput=1

\documentclass[11pt]{article}

\usepackage[final]{acl}

\usepackage{times}
\usepackage{latexsym}

\usepackage[T1]{fontenc}

\usepackage[utf8]{inputenc}

\usepackage{microtype}

\usepackage{inconsolata}

\usepackage{graphicx}

\usepackage{algorithm}
\usepackage{algorithmic}
\usepackage{tcolorbox}
\usepackage{subcaption} 
\usepackage{amsmath}
\usepackage{amssymb}
\usepackage{booktabs} 

\title{Revisiting Chain-of-Thought Prompting: Zero-shot \\ Can Be Stronger than Few-shot}
\author{
Xiang Cheng\textsuperscript{1},
~Chengyan Pan\textsuperscript{1},
~Minjun Zhao\textsuperscript{2},
~Deyang Li\textsuperscript{2},\\
{\bf ~Fangchao Liu\textsuperscript{2},
~Xinyu Zhang\textsuperscript{2},
~Xiao Zhang\textsuperscript{1},
~Yong Liu\textsuperscript{1}\thanks{\ \ Corresponding author.}} \\
\textsuperscript{1}Gaoling School of Artificial Intelligence, Renmin University of China\\
\textsuperscript{2}Huawei Poisson Lab\\
\texttt{\{chengxiang1,2023100836,zhangx89,liuyonggsai\}@ruc.edu.cn}\\
\texttt{\{zhaominjun1,lideyang2,liufangchao,zhangxinyu35\}@huawei.com}
}

\begin{document}
\maketitle
\begin{abstract}
In-Context Learning (ICL) is an essential emergent ability of Large Language Models (LLMs), and recent studies introduce Chain-of-Thought (CoT) to exemplars of ICL to enhance the reasoning capability, especially in mathematics tasks. However, given the continuous advancement of model capabilities, it remains unclear whether CoT exemplars still benefit recent, stronger models in such tasks. Through systematic experiments, we find that for recent strong models such as the Qwen2.5 series, adding traditional CoT exemplars does not improve reasoning performance compared to Zero-Shot CoT. Instead, their primary function is to align the output format with human expectations. We further investigate the effectiveness of enhanced CoT exemplars, constructed using answers from advanced models such as \texttt{Qwen2.5-Max} and \texttt{DeepSeek-R1}. Experimental results indicate that these enhanced exemplars still fail to improve the model's reasoning performance. Further analysis reveals that models tend to ignore the exemplars and focus primarily on the instructions, leading to no observable gain in reasoning ability. Overall, our findings highlight the limitations of the current ICL+CoT framework in mathematical reasoning, calling for a re-examination of the ICL paradigm and the definition of exemplars. Our code is available on GitHub.\footnote{\url{https://github.com/small-xiangcheng/opencompass/tree/my-changes}}

\end{abstract}

\section{Introduction}
\begin{figure}
    \centering
    \includegraphics[width=\linewidth]{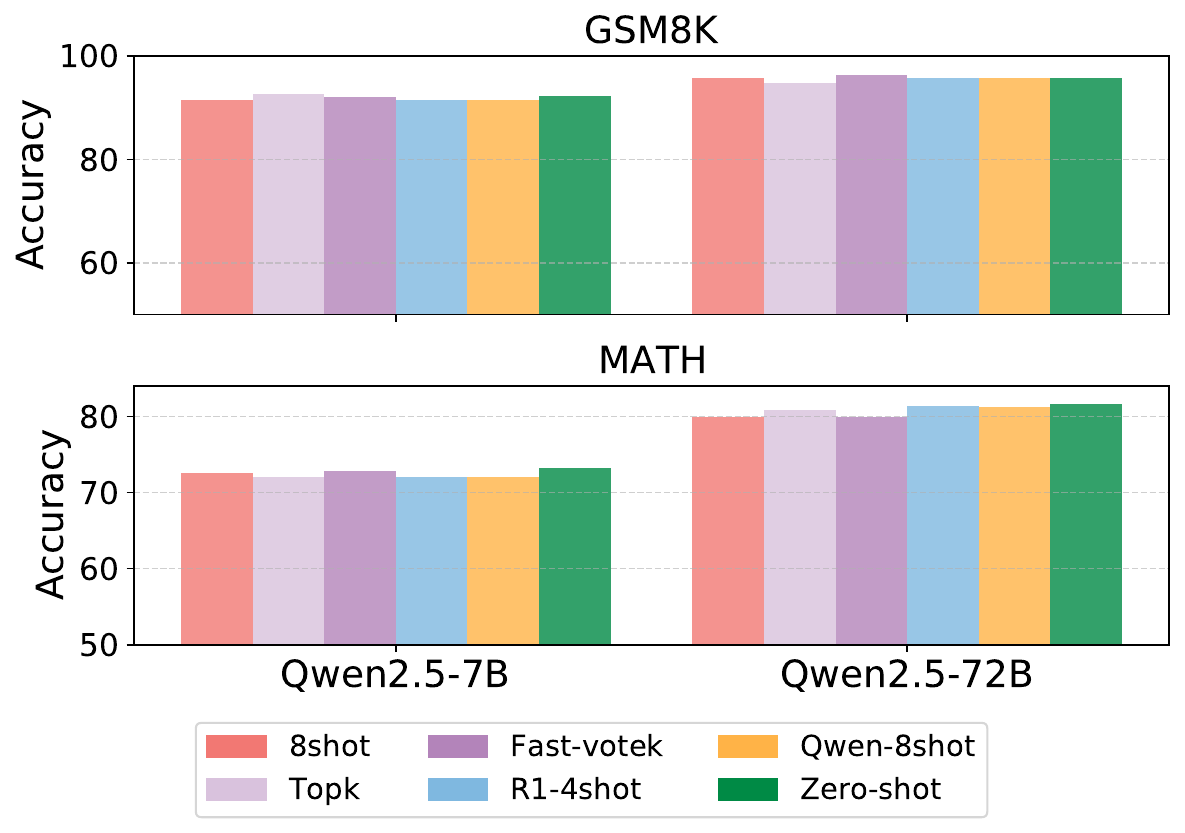}
    \caption{Accuracy under different prompting settings on GSM8K (top) and MATH (bottom). We observe that the Zero-shot setting consistently achieves strong performance, suggesting that the model may not attend to the CoT exemplars. See Section~\ref{sec:exp:exemplars-not-essential} for the full experimental results.}
    \label{fig:intro-findings}
\end{figure}

\begin{figure}
    \centering
    \includegraphics[width=\linewidth]{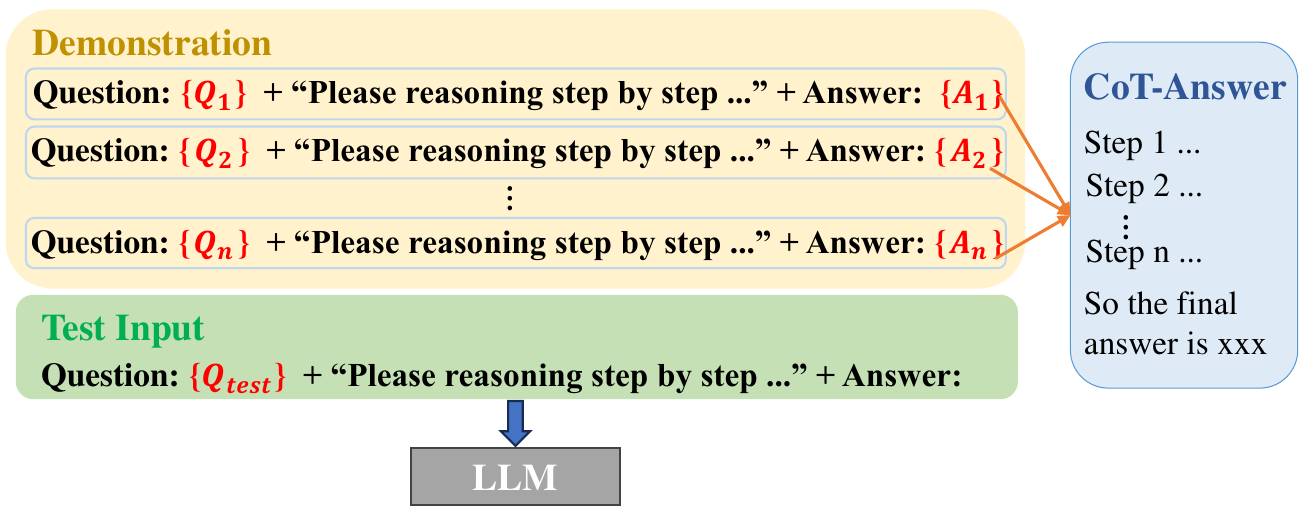}
    \caption{An overview of ICL and CoT prompting. The figure illustrates the Few-shot CoT setting, where the model performs reasoning based on provided demonstrations and a test question. When no demonstrations are given, the setting corresponds to Zero-shot CoT.}
    \label{fig:ICL-COT overview}
\end{figure}

As Large Language Models (LLMs) continues to scale, LLMs exhibit emergent In-Context Learning (ICL) capabilities~\cite{brown-GPT3}, enabling them to perform target tasks by conditioning on a few exemplars without any additional parameter updates. Furthermore, the use of Chain-of-Thought (CoT) exemplars~\cite{wei-COT} in ICL guides models to reason step-by-step. This approach is commonly referred to as \textit{Few-shot CoT}. \citet{kojima-step-by-step} further showed that simply appending the instruction ``Let's think step by step'' can trigger multi-step reasoning even without exemplars, giving rise to the \textit{Zero-shot CoT} paradigm, an overview of them is shown in Figure~\ref{fig:ICL-COT overview}.

Existing research primarily focuses on how the quality, order, and number of exemplars influence ICL performance, proposing various strategies for exemplar construction and selection to enhance model performance across different task settings~\cite{lu-fantastically-order,chen-self-ICL,kim-SG-ICL,purohit-etal-2024-explora}. In addition, several studies have investigated the underlying mechanisms and influencing factors of ICL from either theoretical or empirical perspectives~\cite{Ruifeng-ICL,xie-implicit-Bayesian,Min-Rethinking-demo,wei2023icl-label,wang2023towards}. However, most of these strategies and experimental conclusions are based on earlier, weaker models. As foundation models become increasingly powerful, it is necessary to revisit a central question: \textit{In mathematical reasoning tasks, can CoT exemplars still improve the reasoning performance of recent strong models?}

In this paper, we aim to investigate the actual role of CoT exemplars in mathematical reasoning tasks. We conduct systematic experiments on two representative math reasoning datasets, GSM8K~\cite{cobbe-gsm8k} and MATH~\cite{hendrycks2021math}, using several recent open-source LLMs. We first identify a common evaluation bias in open-source evaluation frameworks~\cite{2023opencompass,open-instruct} in GSM8K, which significantly \textbf{underestimates the performance of Zero-shot CoT}, as discussed in Section~\ref{sec:exp:align-format}. After correcting for this issue, we compare Few-shot CoT with Zero-shot CoT prompting. Our results show that recent strong models already exhibit strong reasoning capabilities under the Zero-shot CoT setting, and the primary role of Few-shot CoT exemplars is to \textbf{align the output format with human expectations}. Subsequent analysis confirms that \textbf{adding traditional CoT exemplars does not improve reasoning performance} (See Section~\ref{sec:exp:traditional-exemplars-not-essential}). Inspired by recent advances in reasoning models with more sophisticated capabilities~\cite{guo2025R1,jaech2024O1}, we then examine the effectiveness of enhanced CoT demonstrations constructed using answers generated by advanced models such as \texttt{Qwen2.5-Max} and \texttt{DeepSeek-R1}. Experimental results indicate that, regardless of enhancement, models tend to ignore the content of exemplars in mathematical reasoning tasks and fail to acquire advanced capabilities such as self-reflection (See Section~\ref{sec:exp:enhanced-exemplars-not-essential}). As a result in figure~\ref{fig:intro-findings}, \textbf{CoT exemplars do not lead to improved reasoning performance in recent models}.

To summarize, our main empirical findings in mathematical reasoning tasks are as follows:
\begin{enumerate}
    \item The primary function of CoT exemplars is to align the output format, and this effect persists regardless of the model's reasoning ability.
    \item Traditional CoT exemplars do not enhance the reasoning performance of strong models, although they may benefit weaker models.
    \item Enhanced CoT exemplars also fail to improve reasoning ability in strong models, as these models tend to ignore the CoT content.
\end{enumerate}

\section{Related Work}
\begin{figure*}
    \centering
    \includegraphics[width=\textwidth]{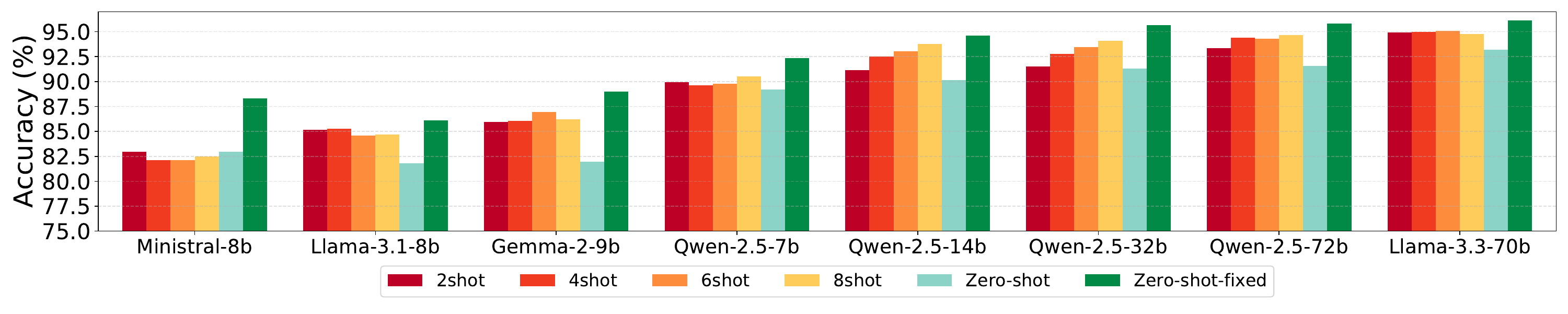}
    \caption{Accuracy of different models on the GSM8K dataset under varying numbers of exemplars. \texttt{Few-shot} examples are taken from \citet{wei-COT}. Only \texttt{Zero-shot-fixed} applies evaluation bias correction, as described in Section~\ref{sec:exp:align-format}; all other settings retain uncorrected results for comparison.}
    \label{fig:gsm-large-align-format}
\end{figure*}
\begin{figure*}
    \centering
    \includegraphics[width=\textwidth]{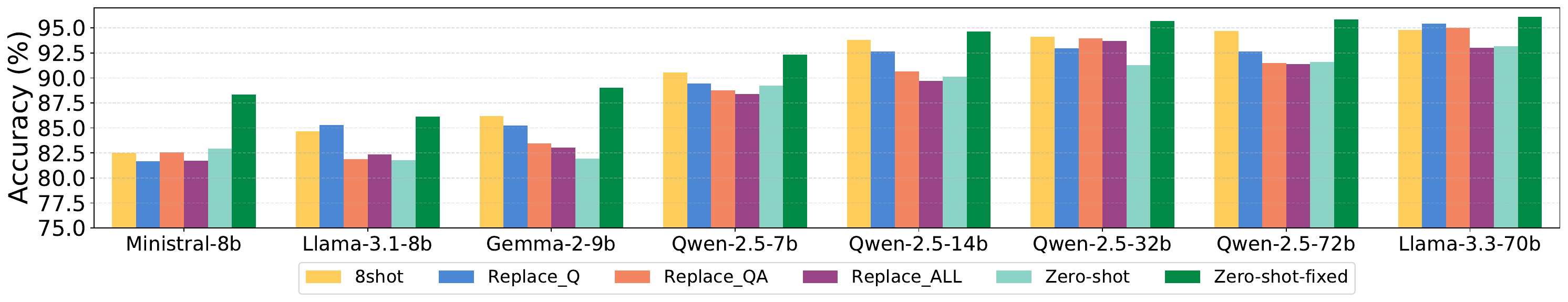}
    \caption{Accuracy of different models on the GSM8K dataset under various ablation settings. \texttt{Replace\_Q} denotes replacing the question in each exemplars with ``xxx''. \texttt{Replace\_QA} replaces both the question and answer with ``xxx'' but retains the final phrase ``So the answer is ...''. \texttt{Replace\_ALL} replaces the question, answer, and the final phrase with ``xxx''. See figure \ref{demo:replace_q}, \ref{demo:replace_qa}, and \ref{demo:replace_all} for input examples, respectively. Other settings follow those in Figure~\ref{fig:gsm-large-align-format}.}

    \label{fig:gsm-large-align-format-ablation}
\end{figure*}

\textbf{CoT Prompting} \quad
ICL enables LLMs to perform tasks without fine-tuning~\cite{brown-GPT3}, but it often falls short in complex reasoning scenarios. To address this, CoT prompting~\cite{wei-COT} introduces intermediate reasoning steps to guide model outputs. Building on CoT, researchers have proposed various extensions to enhance reasoning capabilities. For instance, Tree-of-Thought~\cite{yao-ToT} generalizes CoT to tree-structured reasoning, while Graph-of-Thought~\cite{besta-GoT} further expands it to graph-based structures. The Least-to-Most framework~\cite{zhou-leasttomost} decomposes complex problems into simpler subproblems and solves them sequentially.

\textbf{Exemplar Selection} \quad
In addition to improving CoT itself, numerous studies have explored how exemplar quality, quantity, diversity, and ordering affect ICL performance~\cite{lu-fantastically-order,Li-unified-demonstration,ma2023fairnessguided,zhang-activate-Selection}. A variety of exemplar selection strategies have been proposed. \citet{Fu-Complexity-based} recommend selecting exemplars with higher reasoning complexity (i.e., involving more intermediate steps), while \citet{SU-selective-annotation} emphasize diversity and introduce the VoteK algorithm. Other representative methods include DPP~\cite{ye2023dpp}, a probabilistic model that measures both subset diversity and quality by using the determinant of a kernel matrix—the larger the determinant, the more diverse and representative the selected exemplars. MMR~\cite{ye-etal-2023-mmr} balances relevance and diversity by iteratively selecting items that are highly relevant to the query while remaining dissimilar to already chosen exemplars. EXPLORA~\cite{purohit-etal-2024-explora} evaluates exemplar subsets without relying on model confidence scores.

\textbf{Understanding CoT Prompting} \quad
Beyond methodology, a growing body of research has sought to understand the mechanisms behind ICL and CoT prompting. Theoretical investigations~\cite{dai-meta-gradient,li-softmax-regression,Ruifeng-ICL,mahankali-one-step} offer insights into the learning dynamics of ICL, while empirical studies probe the effectiveness of CoT. For instance, \citet{Min-Rethinking-demo} suggest that exemplars primarily provide distributional rather than semantic information—though their analysis is limited to classification tasks. In the context of reasoning, \citet{levy-ICL-length} report that longer input contexts may hurt performance, and \citet{sprague2025toCoT} find that the benefits of CoT are mainly confined to mathematical and logical reasoning.

Our work complements these lines of research through a systematic empirical study on mathematical reasoning. While prior studies have provided important insights, they are mostly based on earlier and weaker models, whose conclusions may not fully extend to recent, stronger models. We find that, for recent strong models, CoT exemplars primarily function to align output format rather than enhance reasoning ability. This challenges the prevailing assumption that CoT-based ICL reliably improves performance in math reasoning tasks.

\section{Experimental Setup}
\label{sec:experiments-setup}

\textbf{Models} \quad To thoroughly validate our conclusions, we evaluate a variety of open-source language models, including the Qwen2.5 series (ranging from 0.5B to 72B parameters)~\cite{yang2024qwen2.5}, the LLaMA3 series (1B to 70B)~\cite{grattafiori2024llama3}, the Gemma2 series (2B and 9B)~\cite{team2024gemma2}, and Ministral-8B ~\cite{ministral2024}. In addition, to examine the effectiveness of CoT prompting on earlier and weaker models, we include LLaMA2-7B~\cite{touvron2023llama2} and Qwen-7B~\cite{bai2023qwen} for comparative analysis. All models used in our experiments are instruction-tuned variants. More details can be found in Appendix~\ref{sec:appendix:model_details}.

\textbf{Datasets} \quad We focus on mathematical reasoning tasks and conduct experiments on two datasets of varying difficulty: GSM8K~\cite{cobbe-gsm8k} and MATH~\cite{hendrycks2021math}. To ensure accuracy, we perform inference and evaluation on the full test sets of both datasets and report the complete results. More details can be found in Appendix~\ref{sec:appendix:data_details}.

\begin{figure*}
    \centering
    \includegraphics[width=\textwidth]{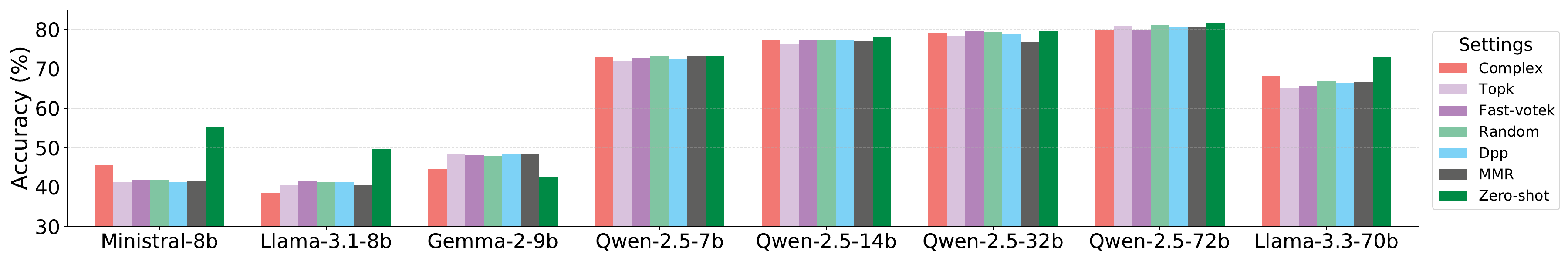}
    \centering
    \includegraphics[width=\textwidth]{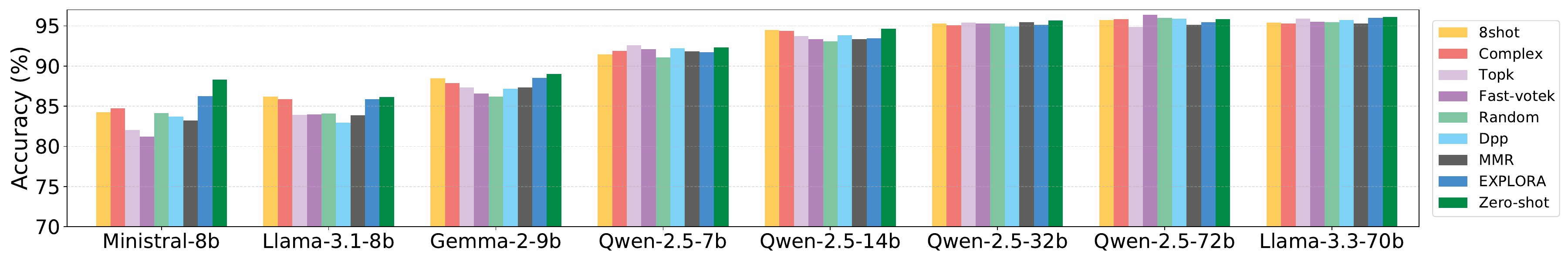}
    \caption{Accuracy of different models under various retrieval methods with a fixed number of 8 retrieved exemplars. The top figure shows results on the MATH dataset, and the bottom figure shows results on the GSM8K dataset.}
    \label{fig:gsm-math-large-retrieval-8shot}
\end{figure*}

\textbf{Environment and Hyperparameters}
We utilize the open-source inference framework OpenCompass~\cite{2023opencompass} and \texttt{vLLM}~\cite{kwon2023vllm} as the backend to run all experiments. Notably, all experiments incorporate a CoT instruction in the prompt: "Please reason step by step, and put your final answer within \textbackslash boxed\{\}." For reproducibility, all experiments are conducted using a fixed random seed of 42. Notably, since greedy decoding is deterministic, the fixed seed does not influence the inference results under a fixed hardware setup. Hence, we do not report the mean or standard deviation of the results. More details can be found in Appendix~\ref{sec:appendix:other_details}.

\section{Exemplars Help Mitigate Evaluation Bias}
\label{sec:exp:align-format}

\textbf{Evaluation Bias in GSM8K} \quad Existing evaluation frameworks for GSM8K (e.g., OpenCompass~\cite{2023opencompass}, Open-Instruct~\cite{open-instruct}) typically extract \textbf{the last number} from model outputs as the predicted answer. However, in Zero-shot CoT prompting, answers are often enclosed in ``\texttt{\textbackslash boxed\{\}}" expressions. This mismatch leads to misjudgments during evaluation, as illustrated in Figure~\ref{tab:error_demo}. To address this, we modify the evaluation script to extract the number inside \texttt{\textbackslash boxed\{\}}, reducing artificially low accuracy caused by output-format misalignment. We consider this a form of evaluation bias that affects fair assessment, either due to oversight or simplification.

\textbf{Exemplars Aid Format Alignment} \quad As shown in Figure~\ref{fig:gsm-large-align-format}, after correcting the evaluation method, the \texttt{Zero\_shot\_fixed} setting yields substantial gains, surpassing all others. This indicates that the original poor performance of \texttt{Zero\_shot} stems not from reasoning limitations, but from output-evaluation mismatch. Moreover, \texttt{Few\_shot} consistently outperforms \texttt{Zero\_shot}, suggesting that exemplars help standardize output format and improve answer extraction. Thus, in math reasoning tasks, the primary benefit of exemplars lies in aligning the model’s output format. Interestingly, for Mistral-8B, exemplars can induce overfitting to simplified reasoning paths, diminishing their effectiveness.

\textbf{Complete Answer Structure is Key Factor} \quad Ablation results in Figure~\ref{fig:gsm-large-align-format-ablation} show a consistent performance drop as more content is masked—from \texttt{Replace\_Q} to \texttt{Replace\_QA} to \texttt{Replace\_All}. This highlights the importance of preserving the full answer structure for effective format alignment. Even partial cues (e.g., “So the answer is \ldots”) prove beneficial, whereas fully removing informative content reverts performance to the \texttt{Zero\_shot} baseline. This confirms that exemplars primarily guide answer formatting rather than reasoning itself.

\section{CoT Exemplars can't improve reasoning ability of strong models}
\label{sec:exp:exemplars-not-essential}

The preceding sections have shown that the primary contribution of exemplars lies in aligning the output format rather than enhancing reasoning ability. However, since we previously used a fixed set of 8 exemplars, an open question remains: \textit{Can exemplars improve the reasoning ability of recent LLMs if we consider different impact factors such as retrieval method, model's intrinsic ability and the quality of exemplars?} 

\begin{figure*}
    \centering
    \includegraphics[width=\textwidth]{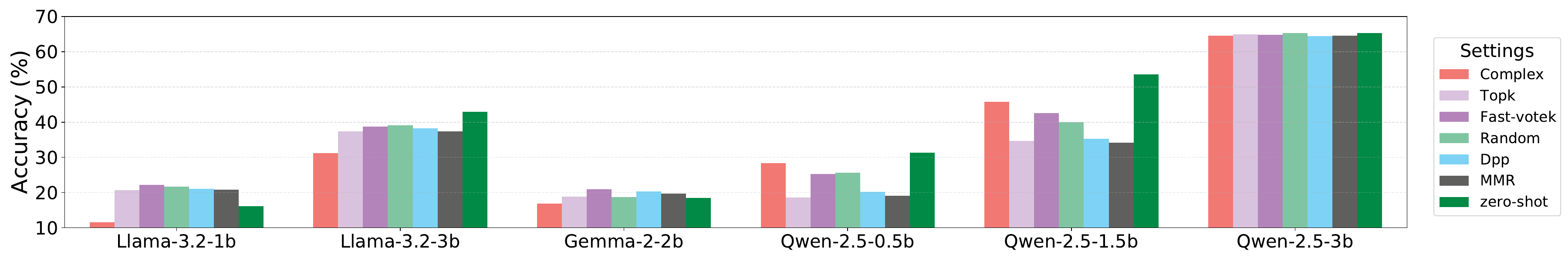}
    \centering
    \includegraphics[width=\textwidth]{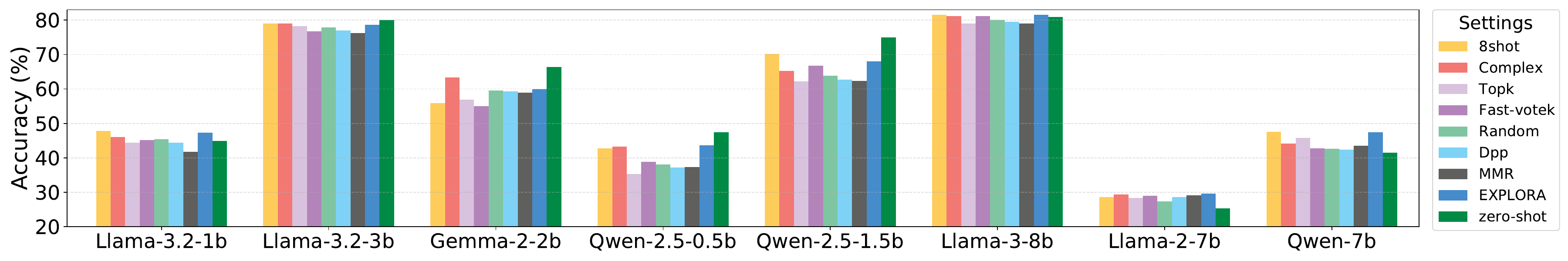}
    \caption{Accuracy of different weaker models under various retrieval methods with a fixed number of 8 retrieved exemplars. The top figure shows results on the MATH dataset, and the bottom figure shows results on the GSM8K dataset.}
    \label{fig:gsm-math-small-retrieval}
\end{figure*}

\subsection{The Impact of the retrieval method}
\label{sec:exp:traditional-exemplars-not-essential}
In this section, we revisit the classical CoT prompting paradigm, in which in-context exemplars are retrieved from the training set of the original dataset. This setup aligns with prior work and allows us to evaluate whether recent large language models (LLMs) still benefit from exemplars under this conventional configuration. To ensure consistency, we apply our corrected evaluation method across a variety of models and compare their performance on GSM8K and MATH using several established exemplar selection strategies. These include Complexity-based selection~\cite{Fu-Complexity-based}, Fast-Votek~\cite{SU-selective-annotation}, DPP~\cite{ye2023dpp}, MMR~\cite{ye-etal-2023-mmr}, and EXPLORA~\cite{purohit-etal-2024-explora}, along with simple TopK and Random baselines.

\begin{figure}
    \centering
    \includegraphics[width=\linewidth]{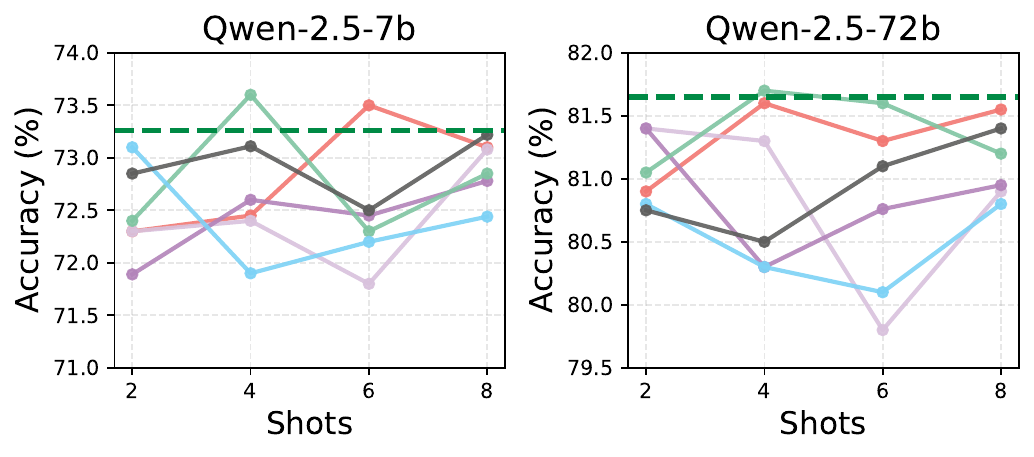}
    \centering
    \includegraphics[width=\linewidth]{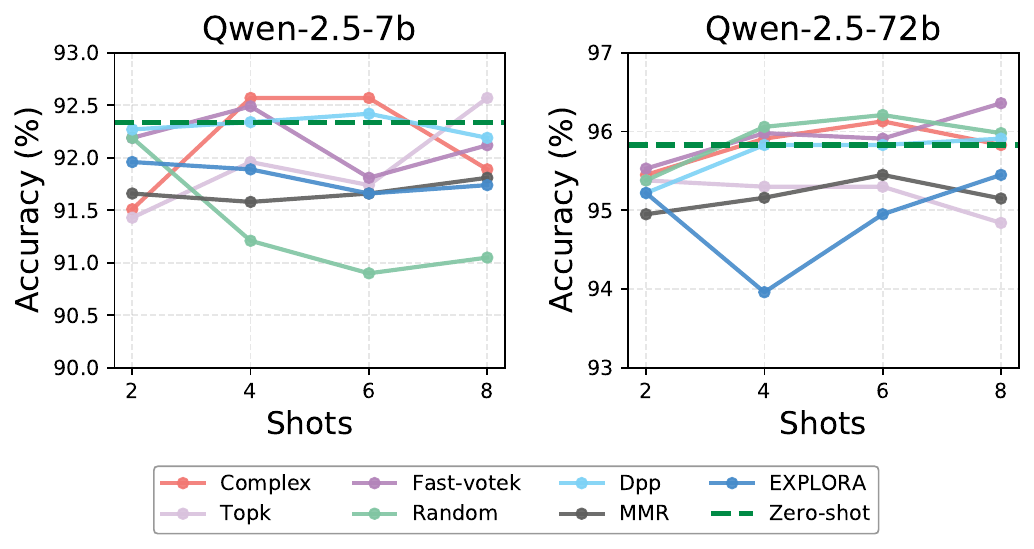}
    \caption{Accuracy variation with different numbers of retrieved exemplars under various retrieval methods, evaluated using Qwen2.5-7B and Qwen2.5-72B. The top figure shows results on the MATH dataset, and the bottom figure shows results on the GSM8K dataset.}
    \label{fig:qwen-gsm-math-retrieval-nshot}
\end{figure}

\textbf{Retrieval-Based Methods Fall Short of Zero-Shot Performance} \quad 
We uniformly retrieve 8 exemplars for each selection method and report the results in Figure~\ref{fig:gsm-math-large-retrieval-8shot}. Across most configurations—regardless of model or dataset— \texttt{Few-shot} performance with retrieval-based methods is comparable to or worse than the \texttt{Zero-shot} baseline. This observation suggests that for advanced language models, in-context exemplars do not enhance reasoning ability, but primarily function is align output formats. Notably, there are a few exceptions. For example, LLaMA3.1-8B exhibits marginal improvements under the 8-shot setting. However, we attribute this to inherent experimental variance rather than genuine reasoning gains. A detailed analysis is provided in Appendix~\ref{app:inherent deviation}.

\textbf{Varying the Number of Exemplars Still Fails to Surpass Zero-Shot} \quad 
Given that using 8 retrieved exemplars often fails to outperform the Zero-shot baseline, we further investigate the impact of varying the number of in-context exemplars. As shown in Figure~\ref{fig:qwen-gsm-math-retrieval-nshot}, Zero-shot prompting achieves the highest accuracy in most settings. Nevertheless, certain retrieval methods occasionally yield slightly better performance, particularly on GSM8K. For example, the Complexity-based retrieval method marginally outperforms Zero-shot when retrieving 4 or 6 exemplars on two different models. However, the improvements are minimal—around 0.2\% in accuracy. It can be reasonably attributed to inherent evaluation variance. Such small fluctuations are more likely to occur on relatively simpler datasets like GSM8K. In contrast, on the more challenging MATH dataset, nearly all retrieval-based configurations consistently underperform relative to the Zero-shot baseline.

Overall, these results reinforce the conclusion that Zero-shot prompting remains the most effective approach in the vast majority of cases. This supports the emerging perspective that traditional CoT prompting paradigms no longer significantly enhance the reasoning capabilities of recent LLMs.

\begin{figure*}
    \centering
    \includegraphics[width=\textwidth]{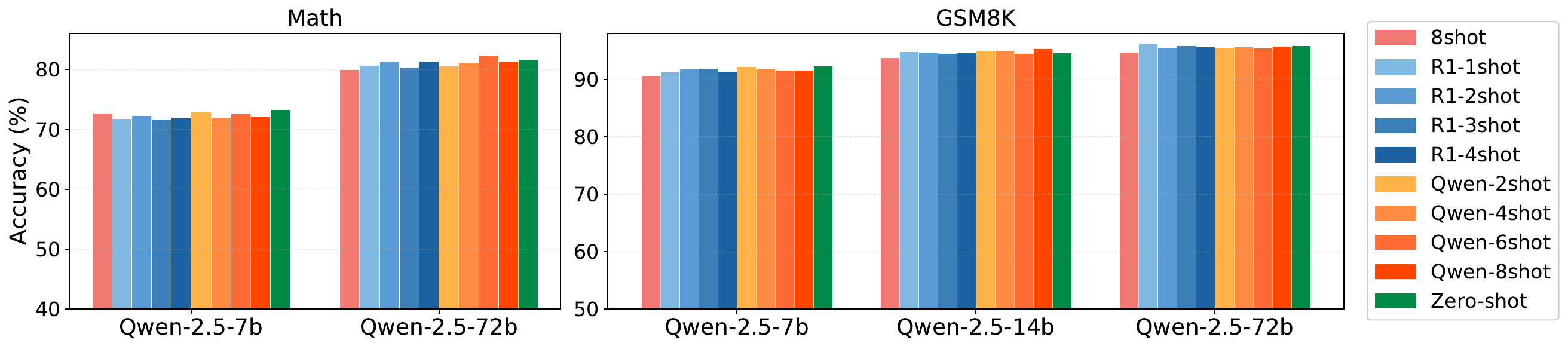}
    \caption{Accuracy under different numbers of exemplars when using DeepSeek R1 responses(marked as R1-nshot) and Qwen2.5-max responses(marked as Qwen-nshot) as exemplars. The left figure shows results on the MATH dataset, and the right figure shows results on the GSM8K dataset.}
    \label{fig:qwen-math-gsm8k-enhance}
\end{figure*}

\subsection{The Impact of exemplars Is Determined by the Model's Intrinsic Capability}

In the previous experiments, we observed that in-context exemplars do not enhance the reasoning ability of recent models such as Qwen2.5 series. \emph{Does this contradict earlier findings from exemplar selection studies, such as those by Fu et al.~\cite{Fu-Complexity-based}?} To further investigate the role of exemplars, we conducted experiments on relatively weaker models. Specifically, we evaluated a set of smaller but recent models (LLaMA3.2-1B, LLaMA3.2-3B, Qwen2.5-0.5B, Qwen2.5-1.5B), as well as several older models (LLaMA3-8B, LLaMA2-7B, Qwen-7B). The same prompt templates were used as in previous experiments, and all model responses were post-processed to eliminate evaluation artifacts and isolate the true effect of exemplars.

Since all outputs were corrected prior to evaluation, the only potential benefit of in-context exemplars in this experiment lies in improving reasoning ability, not output alignment. As shown in Figure~\ref{fig:gsm-math-small-retrieval}, model performance varies significantly. For relatively strong models such as LLaMA3.2-3B and Qwen2.5-1.5B, the \texttt{Zero\_shot} setting yields the highest accuracy, indicating that adding exemplars does not improve reasoning . This is consistent with our findings on stronger models, reaffirming that for capable models, exemplars primarily serve as output format guides rather than improve reasoning. 

However, for weaker models (e.g., LLaMA3.2-1B) and older models with larger parameter counts (e.g., LLaMA2-7B and Qwen-7B), we observe a significant improvement in accuracy when exemplars are provided. This suggests that for such models, in-context exemplars indeed help augment reasoning by supplying intermediate steps that the model struggles to generate on its own. We hypothesize that these weaker or older models lack the complex reasoning patterns that more recent models have acquired through pretraining and instruction tuning, and thus rely more heavily on external exemplars.

Therefore, we conclude that the effectiveness of CoT exemplars depends on the model’s inherent capabilities. Traditional CoT exemplars do not improve the reasoning ability of already-strong models but can play a supportive role for weaker models. Hence, our findings are not in conflict with previous work; rather, they offer a complementary perspective by showing that the utility of exemplars is model-dependent.

\subsection{Is traditional CoT exemplars too easy for strong models?}
\label{sec:exp:enhanced-exemplars-not-essential}
\begin{figure*}
    \centering
    \includegraphics[width=\textwidth]{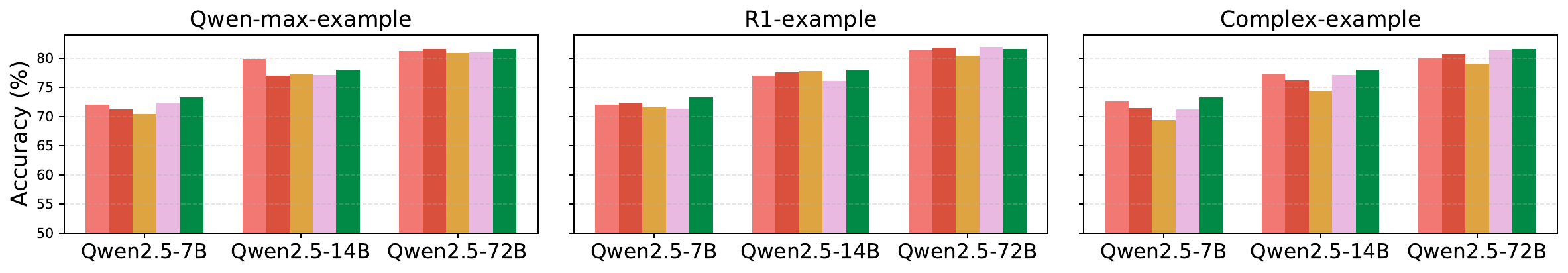}
    \centering
    \includegraphics[width=\textwidth]{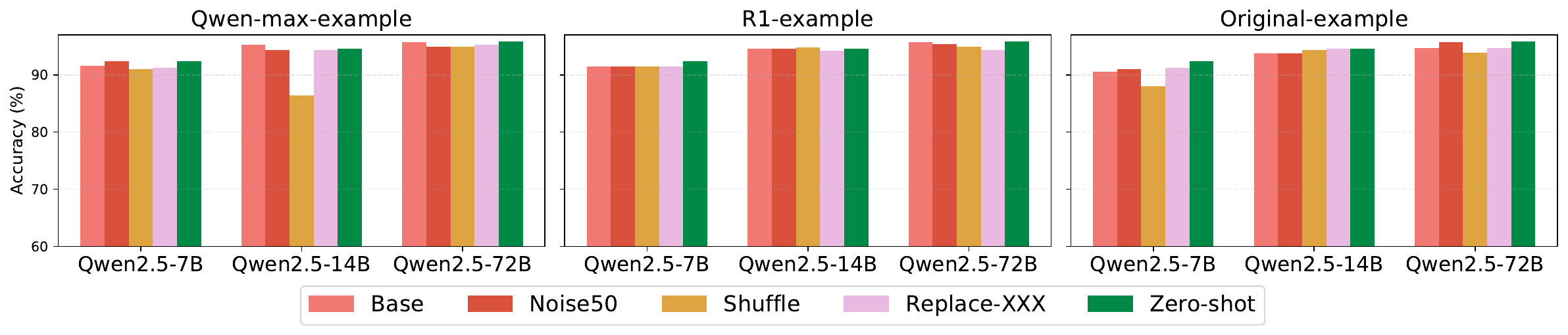}
    \caption{Ablation study on noise injection for three types of exemplars, shown from left to right: exemplars answered by Qwen-max, exemplars answered by R1, and traditional CoT (Chain-of-Thought) exemplars. The top figure shows results on the MATH dataset, and the bottom figure shows results on the GSM8K dataset. \texttt{Base} denotes the original exemplars without noise. \texttt{Noise50} randomly replaces 50\% of the tokens with ``XXX''. \texttt{Shuffle} completely shuffles the words. \texttt{Replace-xxx} replaces all words with ``XXX''.}

    \label{fig:gsm-math-qwen-enhance-ablation}
\end{figure*}

Previous experiments suggest that traditional CoT prompting strategies are largely ineffective for current open-source LLMs. A natural intuition is that the implicit reasoning paths embedded in standard CoT exemplars may be less sophisticated than the models’ own Zero-shot reasoning capabilities. This raises an important question: \textit{can enhanced CoT exemplars benefit these strong models?}

With the emergence of high-performing Reasoning Large Language Models (RLLMs) such as OpenAI o1~\cite{jaech2024O1} and DeepSeek R1~\cite{guo2025R1}, Long Chain of Thought have shown potential in guiding model reasoning. Motivated by this, we consider two enhanced settings: (1) using responses from DeepSeek-R1 as exemplars, and (2) using responses from a stronger LLM, Qwen2.5-Max, as exemplars.  We conduct experiments across the Qwen2.5 family of models (7B, 14B, and 72B). Detailed examples of the input formats are provided in Appendix~\ref{app:input examples}.

\textbf{Quality Helps, but Zero-Shot Still Dominates} \quad 
For each enhanced configuration, we further vary the number of exemplars. Due to the relatively long responses generated by DeepSeek-R1, we accordingly limit the number of exemplars to a maximum of four shots to ensure comparability in input length. The corresponding results are shown in Figure~\ref{fig:qwen-math-gsm8k-enhance}. We observe that enhanced exemplars generally outperform the standard 8-shot CoT setting. In certain configurations, performance may even exceed that of the Zero-shot baseline, such as Qwen2.5-72B on the MATH dataset with the Qwen-6shot setting. Nevertheless, Zero-shot prompting consistently achieves strong accuracy across both datasets without introducing additional context overhead. These findings indicate that while improving exemplar quality is indeed helpful, the reasoning capability of modern large language models is already sufficiently strong that changes in exemplar formatting yield only limited or no improvement over Zero-shot prompting.

\section{Why CoT exemplars is not useful for strong models?}
In this section, we further investigate the reasons behind the ineffectiveness of CoT exemplars. We begin with ablation studies, followed by an analysis of attention visualization results.

\subsection{Ablation Study on Noisy Exemplars}
\label{sec:exp:demo_ablation}
To further investigate why exemplars fail to improve performance, we conduct ablation experiments across three types of CoT exemplars: Traditional CoT, R1-enhanced CoT (from DeepSeek-R1), and Qwen2.5-Max-enhanced CoT. Specifically, for the R1-enhanced configuration, we use 4-shot exemplars, while 8-shot is used for the other settings. We introduce varying levels of noise into the exemplars and evaluate their impact on model performance. Experiments are conducted on the Qwen2.5 series (7B, 14B, and 72B) across both the GSM8K and MATH datasets.

\textbf{Exemplars Are Not Crucial for Recent LLMs} \quad As shown in Figure~\ref{fig:gsm-math-qwen-enhance-ablation}, we observe that in most settings, adding noise to the exemplars does not lead to significant performance degradation. This is especially evident for the larger Qwen2.5-72B model, where even the \texttt{Noise50} configuration can match or slightly outperform the \texttt{Base} setting. These findings suggest that the models may selectively ignore the exemplars and instead rely on their intrinsic reasoning ability. Thus, the performance observed under Few-shot settings may not arise from the informative content of the exemplars, but rather from the model’s inherent Zero-shot capabilities.

\subsection{Attention Visualization}
\label{sec:exp:attention}
\begin{figure}
    \centering
    \begin{subfigure}[b]{0.49\linewidth}
        \centering
        \includegraphics[width=\linewidth]{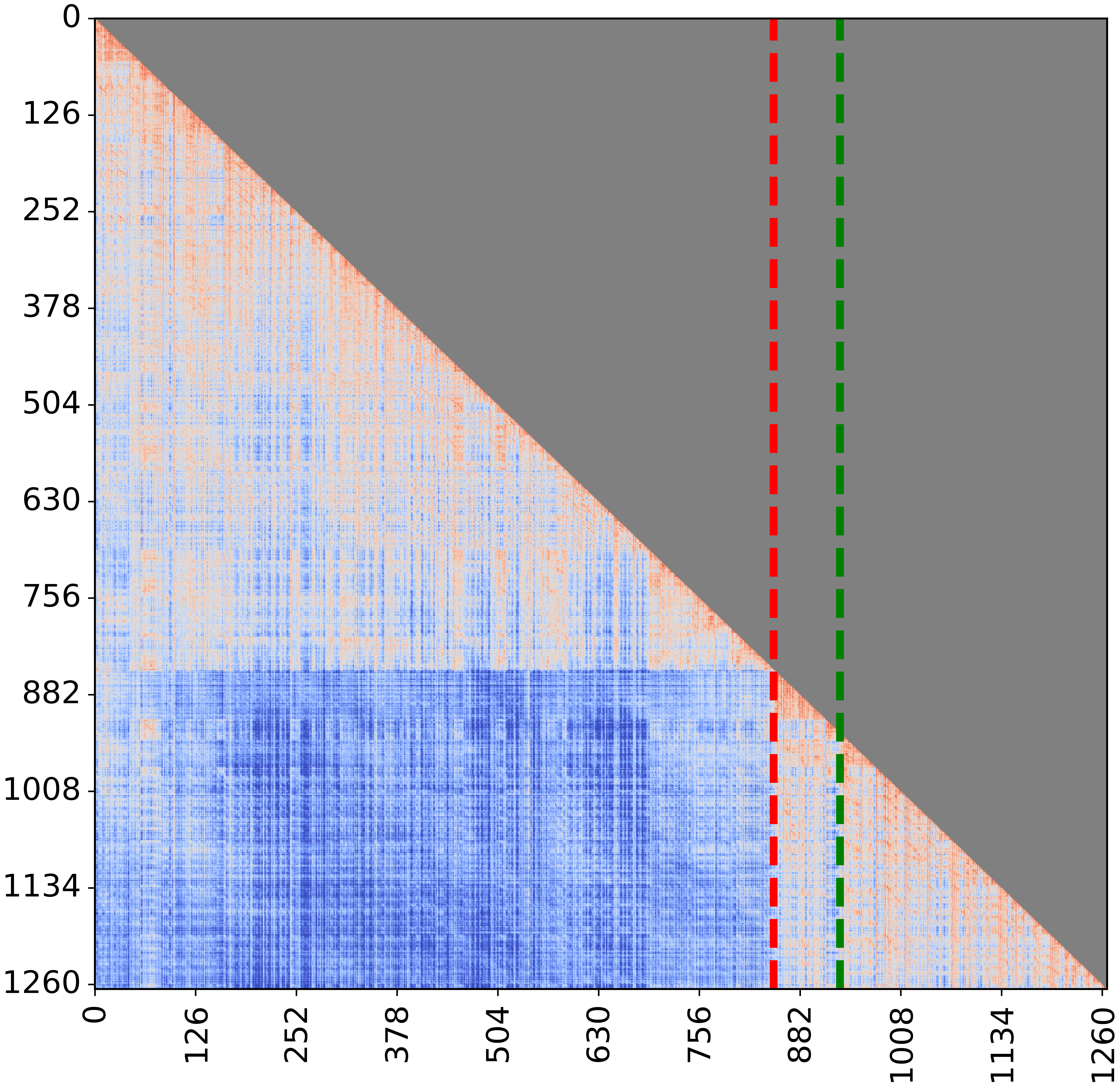}
        \caption{R1-CoT-1shot}
        \label{fig:long_cot_1shot}
    \end{subfigure}
    \hfill
    \begin{subfigure}[b]{0.49\linewidth}
        \centering
        \includegraphics[width=\linewidth]{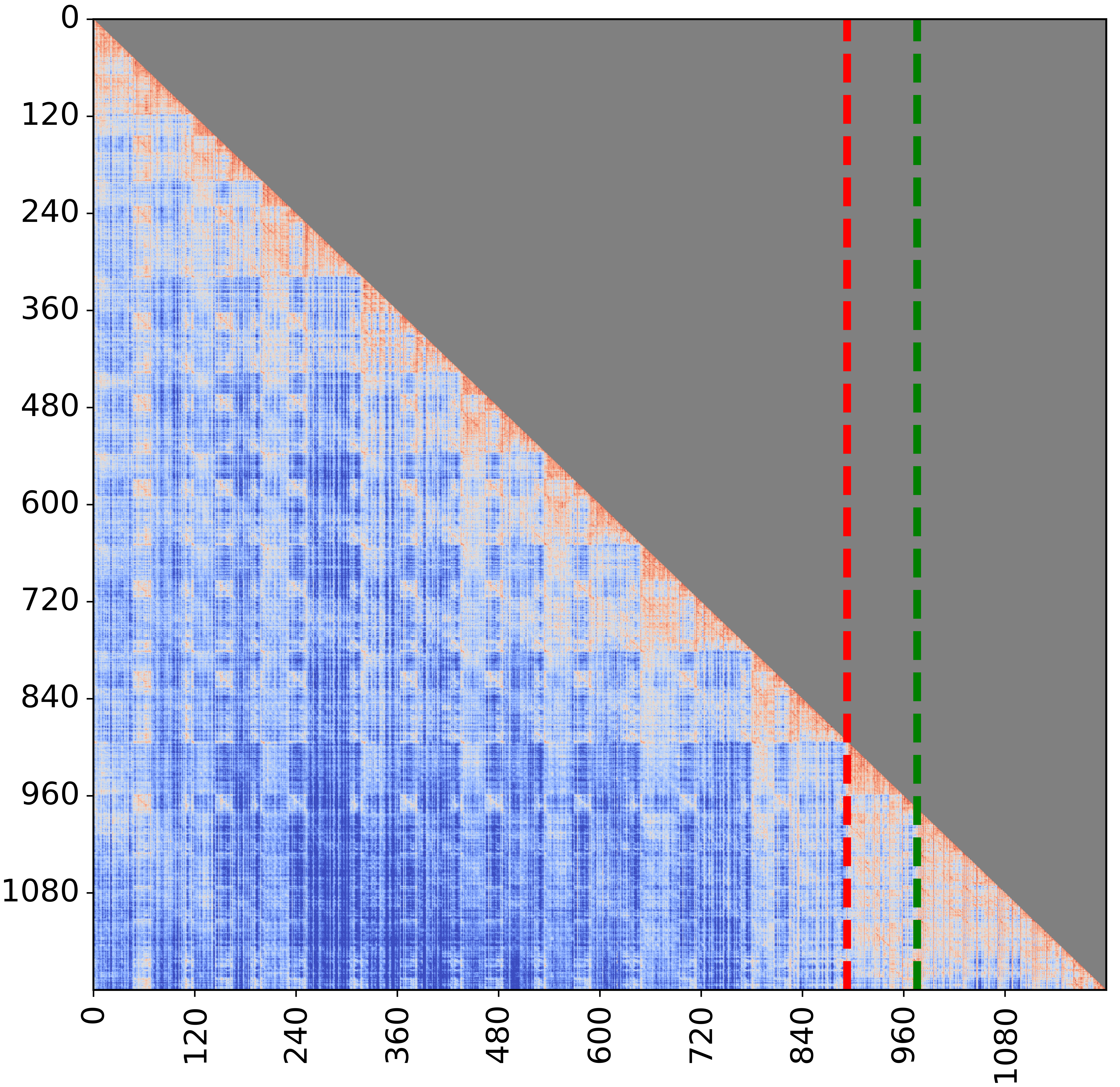}
        \caption{CoT-8shot}
        \label{fig:cot_8shot}
    \end{subfigure}
    
    \vspace{0.4em}

    \begin{subfigure}[b]{0.49\linewidth}
        \centering
        \includegraphics[width=\linewidth]{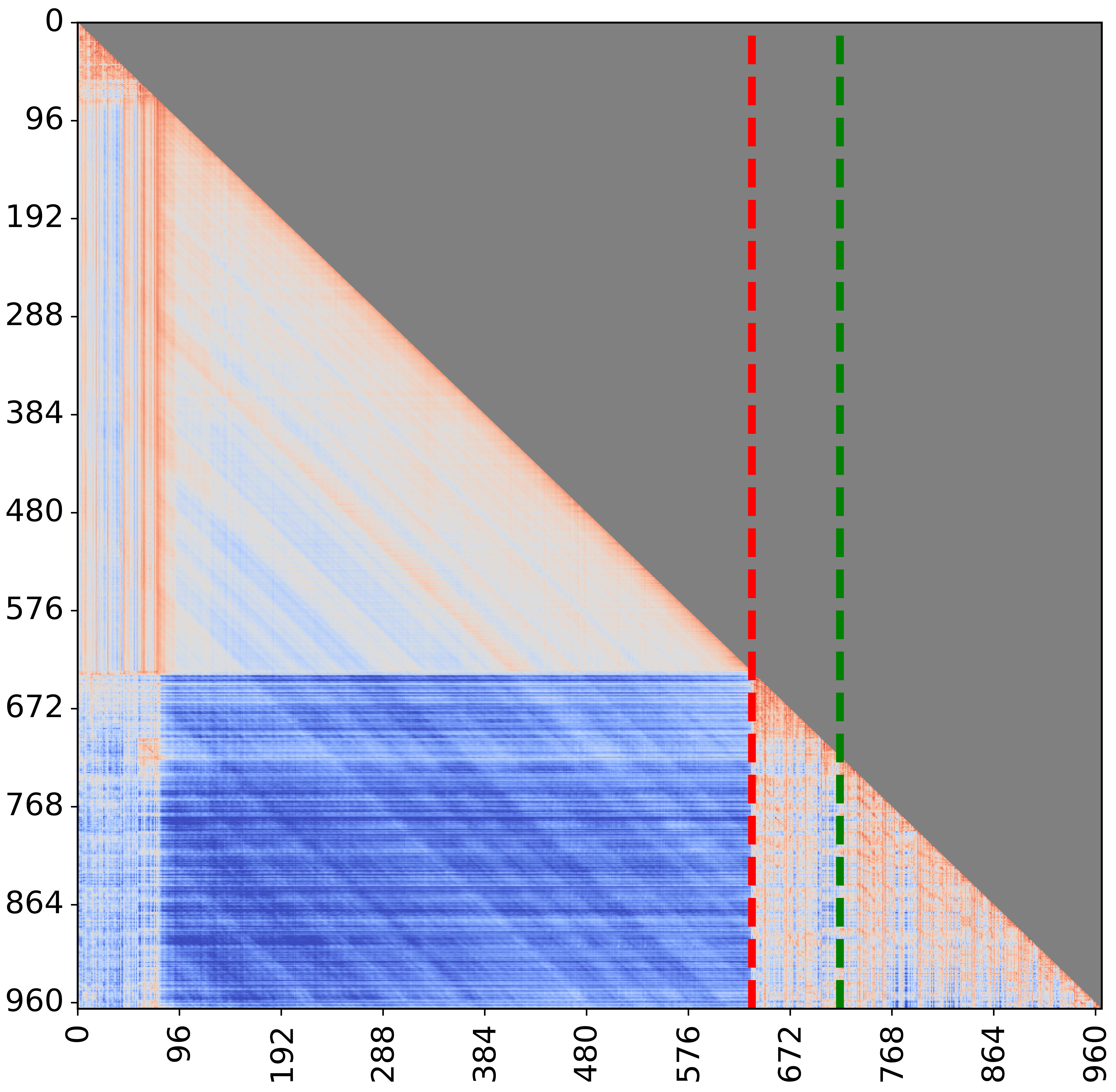}
        \caption{R1-CoT-Replace-XXX}
        \label{fig:long_cot_replace_all}
    \end{subfigure}
    \hfill
    \begin{subfigure}[b]{0.49\linewidth}
        \centering
        \includegraphics[width=\linewidth]{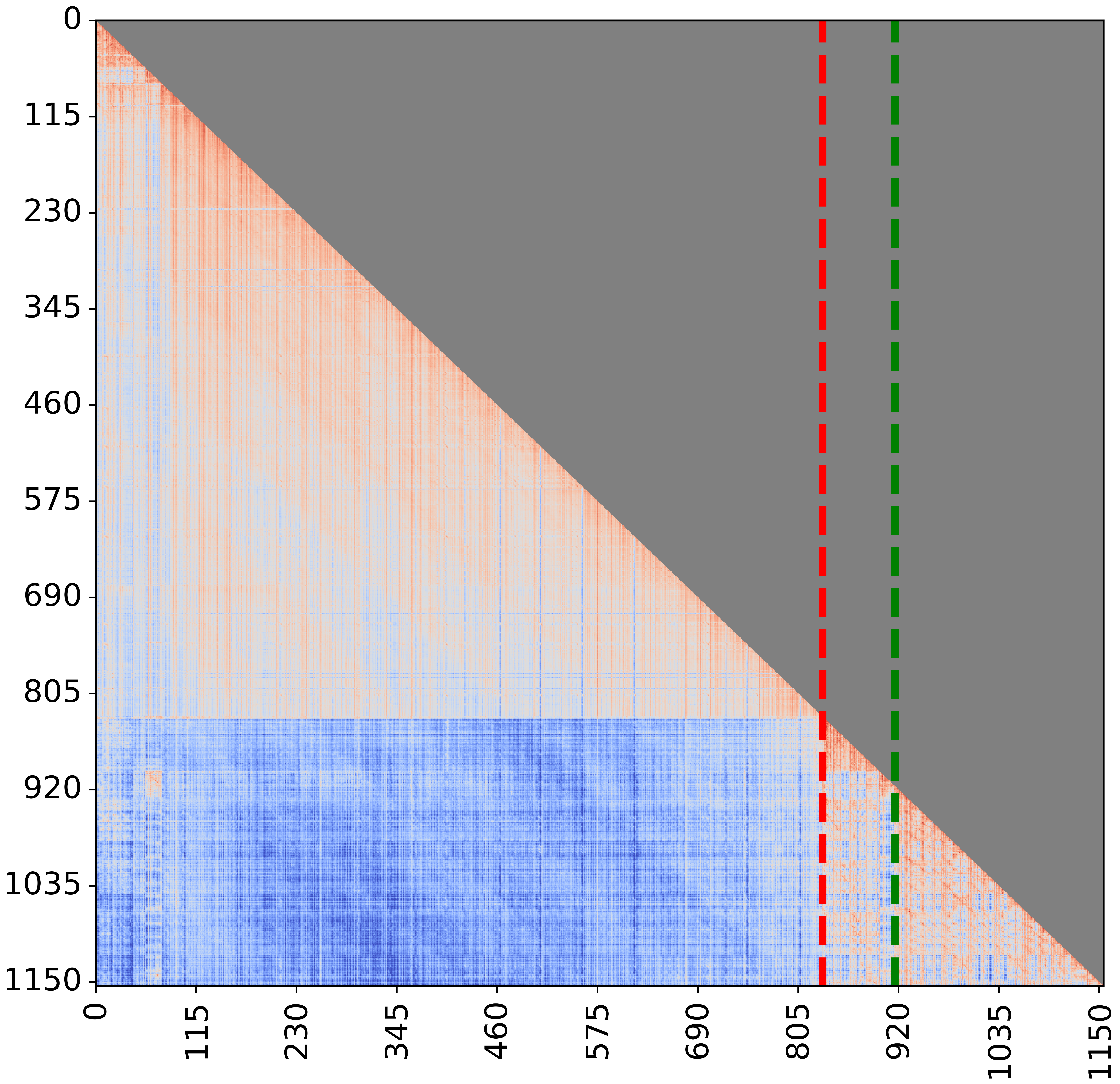}
        \caption{R1-CoT-Shuffle}
        \label{fig:long_cot_shuffle}
    \end{subfigure}

    \caption{Attention visualizations under various settings. The red line indicates the end of the exemplar section, and the green line marks the end of the entire input. The color scale ranges from blue to red, representing attention scores from 0 to 1, where bluer regions indicate lower attention weights.}

    \label{fig:attention_visualization_all}
\end{figure}

The previous results suggest that neither standard CoT prompts nor enhanced exemplars substantially improve model reasoning, and that models may not actively attend to these exemplars during inference. To investigate this further, we analyze the attention distribution of the Qwen2.5-7B model on GSM8K under Few-shot settings. Transformer-based models~\cite{vaswani2017attention} rely on multi-head self-attention, where each head in each layer computes a separate attention matrix. We randomly select a test instance and visualize head 0 in the final (27th) layer. Full visualizations are provided in Appendix~\ref{app:attention visualization}.

As shown in Figure~\ref{fig:attention_visualization_all}, the lower-left region of the attention map—corresponding to the exemplar section—consistently exhibits low scores (blue), while the upper-left region, representing intra-example dependencies, displays stronger attention. The red and green lines mark the ends of the exemplar section and input sequence, respectively; generation begins after the green line.Each attention row reflects how a generated token attends to prior tokens. The weak attention to the exemplars (before the red line) and strong focus on the prompt and test question (between the red and green line) indicate that the model largely ignores exemplars during inference, relying more on the prompt template.

Comparing Figure~\ref{fig:long_cot_1shot} and Figure~\ref{fig:long_cot_replace_all}, we observe slightly higher attention to exemplars in R1-CoT-1shot. However, this does not yield meaningful accuracy gains (see Figure~\ref{fig:gsm-math-qwen-enhance-ablation}), reinforcing that enhanced exemplars have minimal impact on reasoning performance and are largely disregarded by the model.

\section{Discussion and Conclusion}

In this paper, we investigate the role of CoT exemplars in mathematical reasoning tasks with recent strong models within the ICL paradigm. We show that the previously reported low accuracy of the \texttt{zero\_shot} setting stems from limitations in the evaluation script. After correcting the answer extraction process, the \texttt{zero\_shot\_fixed} setting consistently outperforms few-shot CoT prompting. Our findings reveal that: (1) the primary function of exemplars is to align output format; (2) while exemplars benefit weaker models, they fail to enhance the reasoning ability of stronger models. We further explore enhanced CoT exemplar settings and observe moderate improvements over traditional examples. However, ablation studies show that even with noisy or irrelevant exemplars, model accuracy remains stable, indicating that: (3) strong models rely more on prompt templates than on exemplar content. Finally, attention visualizations support this conclusion by demonstrating weak attention to exemplar tokens. Overall, our study highlights the limitations of current ICL+CoT paradigm in mathematical reasoning and calls for a reevaluation of the role of CoT exemplars. We hope our work offers new insights and empirical grounding for future research.

\textbf{Are Existing Evaluation Frameworks Reliable?} \quad  
As discussed in Section~\ref{sec:exp:align-format}, OpenCompass~\cite{2023opencompass} evaluates GSM8K performance by extracting only the final digit from model outputs. This evaluation imposes strict constraints on output format, potentially overlooking genuine reasoning ability. While such an evaluation may be suitable for measuring output format consistency, it can misrepresent a model’s reasoning capabilities. Hence, if the research goal is to evaluate reasoning rather than formatting, care must be taken to avoid evaluation-induced bias. We advocate that future studies place particular emphasis on the potential bias introduced by evaluation frameworks and carefully design experiments to ensure faithful assessment of model behavior.

\textbf{Why Does CoT Prompting Fail for Strong Models?} \quad  
As shown in Section~\ref{sec:exp:demo_ablation}, injecting various levels of noise into exemplars does not significantly degrade performance. Furthermore, attention visualization in Section~\ref{sec:exp:attention} reveals that models allocate minimal attention to the exemplar region. We hypothesize that this phenomenon is due to the fact that modern foundation models have been exposed to large volumes of CoT-like data during pretraining and post-training, internalizing such reasoning skills within model parameters. Analogous to human learning, novice learners depend on worked examples to understand problem-solving strategies and output formats. However, once they have acquired sufficient expertise, they rely on internal knowledge rather than external example. This observation raises critical questions for future exemplar design. For example, what role should exemplars play for RLLMs? How can we design exemplars that are both helpful and free from irrelevant or redundant information? Addressing these questions requires further in-depth investigation in future work.

\textbf{Is Few-shot CoT Still Widely Used in Current LLMs Evaluations?}  
We note that according to DeepSeek's reports, few-shot prompting may lead to performance degradation in DeepSeek-R1 \cite{guo2025R1}. However, our experiments are conducted on \emph{instruction-tuned} models, which differ from RLLMs like DeepSeek-R1 that are typically recommended for zero-shot settings. These RLLMs are known for strong reasoning abilities and are less sensitive to exemplars. Moreover, few-shot evaluation remains an active area of research and practice. Numerous exemplar selection methods~\cite{purohit-etal-2024-explora,ye-etal-2023-mmr,ye2023dpp} continue to focus on identifying the most effective exemplars to enhance model performance. In addition, prominent model families such as Gemma3~\cite{team2025gemma3} and Qwen2.5-Math~\cite{yang2024qwen2-math} are still evaluated using Few-shot CoT. These observations indicate that few-shot CoT remains relevant to both practitioners and researchers. Our results further clarify the conditions under which few-shot CoT is most beneficial, such as with older model versions (e.g., Qwen-7B, now obsolete) and with the latest but smaller-scale models (e.g., LLaMA3.2-1B), which still rely on external exemplars—suggesting its potential for edge-device applications.

\section*{Limitations}

This study looks at CoT prompting for mathematical reasoning. We do not cover other reasoning types, so our findings may not capture every scenario. Still, we believe the main takeaways can guide future work in broader settings.

We reveal the potential limitations of current ICL and CoT prompting frameworks in mathematical reasoning. Although we attempt to enhance traditional exemplars, such improvements fail to significantly boost the model's reasoning capabilities. As such, we do not propose specific solutions to this issue. Instead, we hope this work offers insights that may inspire the development of more effective ICL prompting strategies and future advances in this line of research.

\section*{Acknowledgements}
This research was supported by Huawei Poisson Lab, National Key Research and Development Program of China (NO. 2024YFE0203200), National Natural Science Foundation of China (No.62476277), CCF-ALIMAMA TECH Kangaroo Fund(No.CCF-ALIMAMA OF 2024008), and Huawei-Renmin University joint program on Information Retrieval. We also acknowledge the support provided by the fund for building worldclass universities (disciplines) of Renmin University of China and by the funds from Beijing Key Laboratory of Big Data Management and Analysis Methods, Gaoling School of Artificial Intelligence, Renmin University of China, from Engineering Research Center of Next-Generation Intelligent Search and Recommendation, Ministry of Education, from Intelligent Social Governance Interdisciplinary Platform, Major Innovation \& Planning Interdisciplinary Platform for the “DoubleFirst Class” Initiative, Renmin University of China, from Public Policy and Decision-making Research Lab of Renmin University of China, and from Public Computing Cloud, Renmin University of China.

\bibliography{custom}

\appendix

\section{Experimental Details}
\label{sec:appendix:experimental_details}

\subsection{Model Details}
\label{sec:appendix:model_details}
All models used in our experiments are instruction-tuned variants. Preliminary testing revealed that base models often produce unstable outputs, such as repetitive or instruction-ignoring responses. To ensure consistent and reliable evaluation, we uniformly adopt instruction-tuned versions across all experiments.

\subsection{Data Details}
\label{sec:appendix:data_details}
We evaluate models on two mathematical reasoning benchmarks of varying difficulty: GSM8K~\cite{cobbe-gsm8k} and MATH~\cite{hendrycks2021math}. GSM8K contains 1,319 grade-school word problems, typically requiring 3–4 simple reasoning steps. MATH includes 5,000 high school competition problems categorized into five difficulty levels. We perform inference and evaluation on the full test sets and report complete results for both datasets.

\subsection{Implementation Details}
\label{sec:appendix:other_details}
All experiments are conducted using the OpenCompass~\cite{2023opencompass} framework. We employ \texttt{vLLM}~\cite{kwon2023vllm} as the backend to enable efficient, parallelized inference without sacrificing accuracy. Unless otherwise specified, all prompts include the instruction: \textit{“Please reason step by step, and put your final answer within \textbackslash boxed\{\}.”} For reproducibility purposes, every experiment was performed using a predetermined random seed (42). It's important to highlight that because greedy decoding produces deterministic outputs, this fixed seed has no impact on inference results given identical hardware configurations. Consequently, we have not included statistical measures (e.g., mean or standard deviation) in our results reporting.

\section{Additional Discussion}
\subsection{Comparison Between Base Models and Instruction-Tuned Models}

In the early stages of this work, we experimented with \emph{base models}, but observed significant output instability—such as meaningless repetitions or unsolicited follow-up queries, especially for smaller models (1B–14B). We were concerned that these issues could undermine the robustness of our conclusions, so we chose to perform our main experiments using \emph{instruction-tuned} models, as described in Appendix\ref{sec:appendix:model_details}.

Nevertheless, we provide the evaluation results for the Qwen2.5-72B base model (which may still exhibit slight instability) in Table~\ref{tab:appendix:base_model_results}. These results show that base models still exhibit evaluation bias and rely more heavily on external exemplars, which is consistent with our explanation in the main text.  

\begin{table*}[t]
\centering
\begin{tabular}{lcc}
\toprule
Method & Qwen2.5-72B-Instruct & Qwen2.5-72B-Base \\
\midrule
GSM8K (8-shot)       & 95.75 & 90.52 \\
GSM8K (Zero-shot)    & 91.58 & 84.69 \\
GSM8K (Zero-shot-Fix)& 95.83 & 90.27 \\
Math (8-shot)        & 81.30 & 75.82 \\
Math (Zero-shot)     & 81.64 & 70.60 \\
\bottomrule
\end{tabular}
\caption{Evaluation results for the Qwen2.5-72B base model.}
\label{tab:appendix:base_model_results}
\end{table*}

Importantly, our paper does not define model strength purely by the degree of instruction tuning (which is also correlated with parameter count). Instead, we use a model’s \emph{Zero-shot CoT capability} as an intuitive indicator of its “strength” in our context. Models whose Zero-shot CoT performance surpasses their Few-shot CoT performance are regarded as \emph{strong models}, as they do not depend on external exemplars.  

We emphasize that introducing the concepts of strong and weak models serves to reconcile our findings with prior studies. Earlier work often found Few-shot CoT effective because weaker models, those with insufficient Zero-shot CoT ability were used. As model capabilities improve over time, whether for base or instruction-tuned models, strong models no longer require exemplars for competitive performance.

\subsection{Additional Experiments on More Datasets}

Our paper primarily investigates whether exemplars can enhance reasoning ability, focusing on mathematical reasoning tasks. Many exemplar selection methods have already been evaluated on GSM8K and MATH, and our work complements these studies by more precisely defining the effective range of Few-shot CoT and showing that Zero-shot CoT may be more suitable for \emph{strong models}. Recognizing the importance of out-of-distribution generalization, we also conducted experiments on LSAT~\cite{zhong-etal-2024-agieval} and show results in Table~\ref{tab:appendix:lsat_results}.    Unfortunately, we found that adding exemplars caused severe performance degradation, with models producing repetitive or illogical outputs. This effect was more pronounced in smaller models, further confirming the stability and superiority of Zero-shot CoT. 
\begin{table*}[t]
\centering
\begin{tabular}{lccc}
\toprule
LSAT Task & Qwen2.5-7B-it & Qwen2.5-32B-it & Qwen2.5-72B-it \\
\midrule
LSAT-AR (Few-shot) & --    & 6.96  & 15.62 \\
LSAT-AR (Zero-shot) & 23.91 & 28.70 & 31.84 \\
LSAT-LR (Few-shot) & 16.08 & 8.63  & 41.55 \\
LSAT-LR (Zero-shot) & 60.08 & 83.73 & 84.65 \\
LSAT-RC (Few-shot) & 3.72  & 15.99 & 36.92 \\
LSAT-RC (Zero-shot) & 71.75 & 83.64 & 84.24 \\
\bottomrule
\end{tabular}
\caption{Performance on LSAT benchmarks.}
\label{tab:appendix:lsat_results}
\end{table*} 

We additionally tested other reasoning benchmarks, CommonsenseQA~\cite{talmor-etal-2019-commonsenseqa} and LogicQA~\cite{liu2020logiqa}, comparing only Few-shot and Zero-shot performance. The results are shown in Table~\ref{tab:commonsense_logicqa}.

These results are consistent with our main findings: Zero-shot CoT consistently outperforms Few-shot CoT. We observed that adding exemplars introduces disruptions such as repetition and logical errors, especially in smaller models. Together, these findings indicate that Zero-shot CoT is more stable and reliable—particularly in resource-constrained scenarios where models have fewer than 70B parameters.
\begin{table*}[t]
\centering
\begin{tabular}{lccc}
\toprule
Task & Qwen2.5-7B-it & Qwen2.5-32B-it & Qwen2.5-72B-it \\
\midrule
CommonsenseQA (7-shot)   & 25.14 & 48.57 & 81.24 \\
CommonsenseQA (Zero-shot)& 79.85 & 84.60 & 84.93 \\
LogicQA (3-shot)         & --    & --    & 12.66 \\
LogicQA (Zero-shot)      & 49.31 & 57.30 & 62.96 \\
\bottomrule
\end{tabular}
\caption{Performance on CommonsenseQA and LogicQA benchmarks.}
\label{tab:commonsense_logicqa}
\end{table*}

\subsection{Discussion of Other Potential Evaluation Biases}

\begin{table*}[t]
    \centering
    \begin{tabular}{lccc}
    \toprule
    Model Variant & Qwen2.5-7B-it & Qwen2.5-32B-it & Qwen2.5-72B-it \\
    \midrule
    Zero-shot             & 2 & 1 & 1 \\
    8-shot (Original)     & 1 & 2 & 0 \\
    8-shot (Qwen-Enhanced)& 2 & 0 & 1 \\
    4-shot (R1-Enhanced)  & 0 & 1 & 2 \\
    \bottomrule
    \end{tabular}
    \caption{Number of truncated responses for each model variant.}
    \label{tab:appendix:truncated_responses}
\end{table*}

In addition to the evaluation biases discussed in the main text, real-world evaluations may involve other sources of error. A common issue is \emph{output truncation}, which we discuss further in this section.  

We set the maximum token limit to 4096 and manually inspected model outputs, applying stopwords such as "</s>", "<|im\_end|>", "<|endoftext|>", "\textbackslash n\textbackslash nQuestion:", "<|end\_of\_text|>", "Question", and "\textbackslash n[Question]" to mitigate truncation or formatting problems. For base models, we also observed repeated or fabricated outputs; these issues were largely minimized by using more stable instruction-tuned models.  

To quantify this effect, we manually reviewed the GSM8K dataset and counted evaluation errors caused by overly long responses (e.g., meaningless repetition or excessively long reasoning chains). Among the 1,319 test questions, we identified approximately 100 erroneous responses. Given the dataset size, we consider manual inspection acceptable for gaining accurate insights. Table~\ref{tab:appendix:truncated_responses} summarizes the number of errors caused by overly long outputs. These abnormal cases represent only a very small fraction of the total responses (well under 1,319), and thus do not affect our main conclusions.  

During manual inspection, we also found a handful of other rare error types (typically one or two instances each), such as unit conversion mismatches. For example, in some cases the reference answer was "1 dollar" while the model responded with "100 cents"; similarly, for a time-related query, the reference answer was "2" but the model output "2:00". These examples underscore the importance of considering semantic and mathematical equivalence in evaluation processes. Although such errors are infrequent and do not affect our overall conclusions.

\section{Inherent Biases in the Experiments}
\label{app:inherent deviation}

In this section, we examine inherent biases in the experimental process. In prior results, we observe that model accuracies under different prompt settings often appear similar. For instance, on GSM8K, LLaMA3.1-8B achieves nearly identical performance in both Zero-shot and 8-shot settings (see Figure~\ref{fig:gsm-math-large-retrieval-8shot}).

\begin{figure*}
    \centering
    \begin{subfigure}{0.45\textwidth}
        \centering
        \fbox{\includegraphics[width=\linewidth]{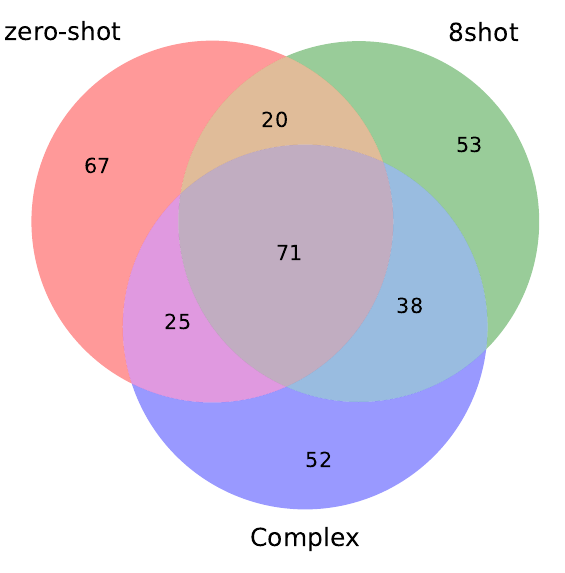}}
        \caption{Zero-shot, 8-shot, and Complex}
    \end{subfigure}
    \hfill
    \begin{subfigure}{0.46\textwidth}
        \centering
        \fbox{\includegraphics[width=\linewidth]{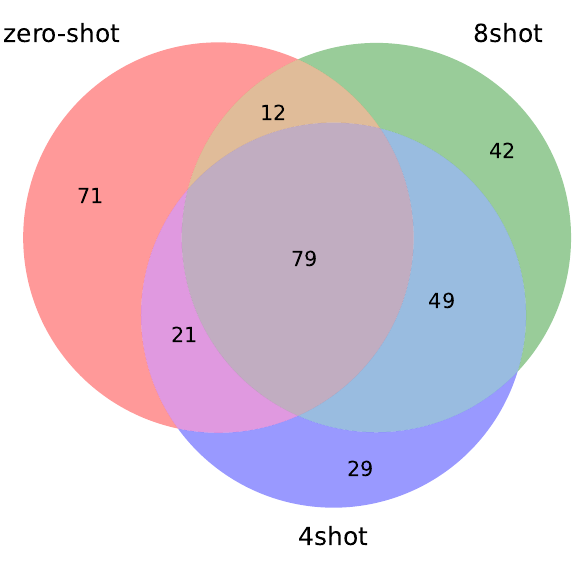}}
        \caption{Zero-shot, 8-shot, and 4-shot}
    \end{subfigure}
    \caption{Error distributions for LLaMA3.1-8B under different prompt settings.}
    \label{fig:llama-3.1-8b}
\end{figure*}

\begin{figure*}
    \centering
    \begin{subfigure}{0.45\textwidth}
        \centering
        \fbox{\includegraphics[width=\linewidth]{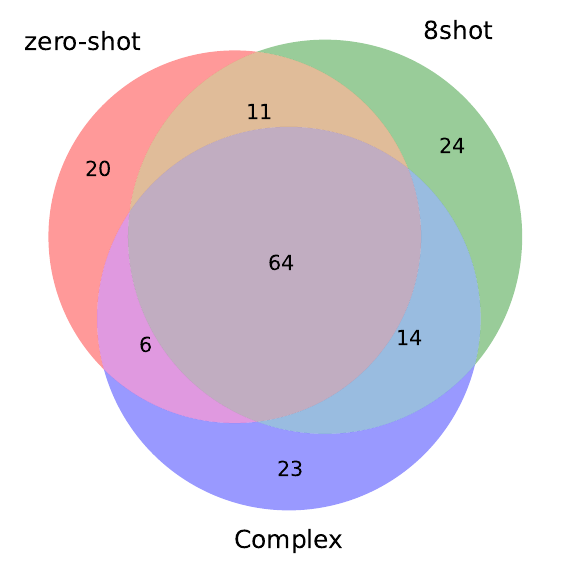}}
        \caption{Zero-shot, 8-shot, and Complex}
    \end{subfigure}
    \hfill
    \begin{subfigure}{0.45\textwidth}
        \centering
        \fbox{\includegraphics[width=\linewidth]{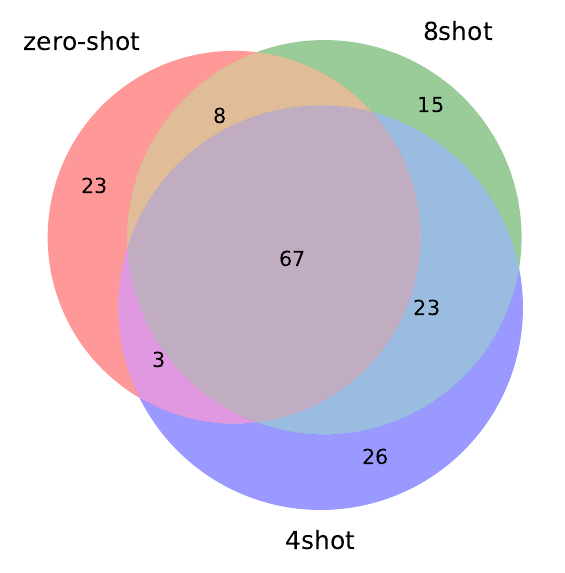}}
        \caption{Zero-shot, 8-shot, and 4-shot}
    \end{subfigure}
    \caption{Error distributions for Qwen2.5-7B under different prompt settings.}
    \label{fig:qwen-2.5-7b}
\end{figure*}

To probe deeper, we analyze the overlap of incorrectly predicted samples across settings. As shown in Figure~\ref{fig:llama-3.1-8b}, while overall accuracy is similar, the error overlap is limited: only 91 shared errors, with 92 unique to Zero-shot and 91 unique to 8-shot. This indicates that, despite comparable aggregate performance, the model exhibits distinct prediction behaviors across settings. Similar patterns hold for other models (see Figure~\ref{fig:qwen-2.5-7b}), suggesting a non-trivial divergence in error distributions.

We attribute this to variation in in-context exemplars, which can subtly influence the model’s internal activations and reasoning paths—introducing an \textbf{inherent bias}. Such biases are widespread and hard to eliminate entirely. Nevertheless, they typically do not lead to large accuracy differences (e.g., a one-sample gap in the above case), implying that aggregate accuracy remains a valid metric for evaluating prompt effectiveness and influence.

\section{Details of Attention Visualization and Supplementary Results}
\label{app:attention visualization}

This section details the attention visualization procedure and presents additional attention distribution results.

Directly visualizing raw attention matrices from the \texttt{transformers} interface~\cite{wolf-etal-2020-transformers} often produces distorted outputs due to extreme value ranges, obscuring actual attention patterns. To mitigate this, we apply a normalization pipeline as described in Algorithm~\ref{alg:scale_attention}.

Specifically, we add a small constant $\epsilon$ for numerical stability, apply a logarithmic transformation to compress the dynamic range, clip values to $[-\tau, 0]$ to suppress outliers, and finally normalize to the $[0,1]$ interval. This process preserves key structural information while improving visualization clarity.

\begin{figure*}
    \centering
    \begin{subfigure}{0.49\textwidth}
        \centering
        \includegraphics[width=\linewidth]{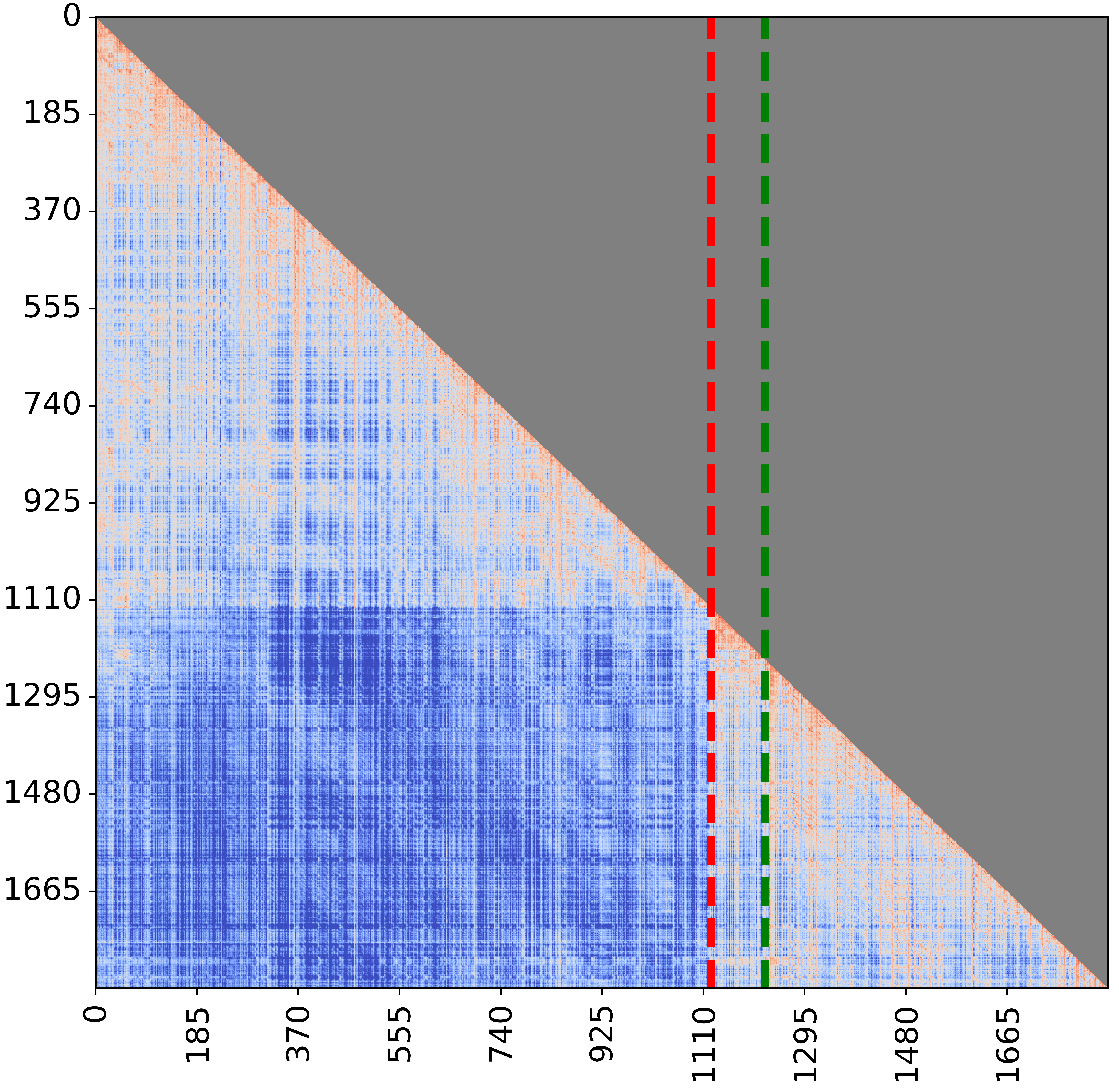}
        \caption{Attention visualization for R1-CoT-1shot}
    \end{subfigure}
    \hfill
    \begin{subfigure}{0.49\textwidth}
        \centering
        \includegraphics[width=\linewidth]{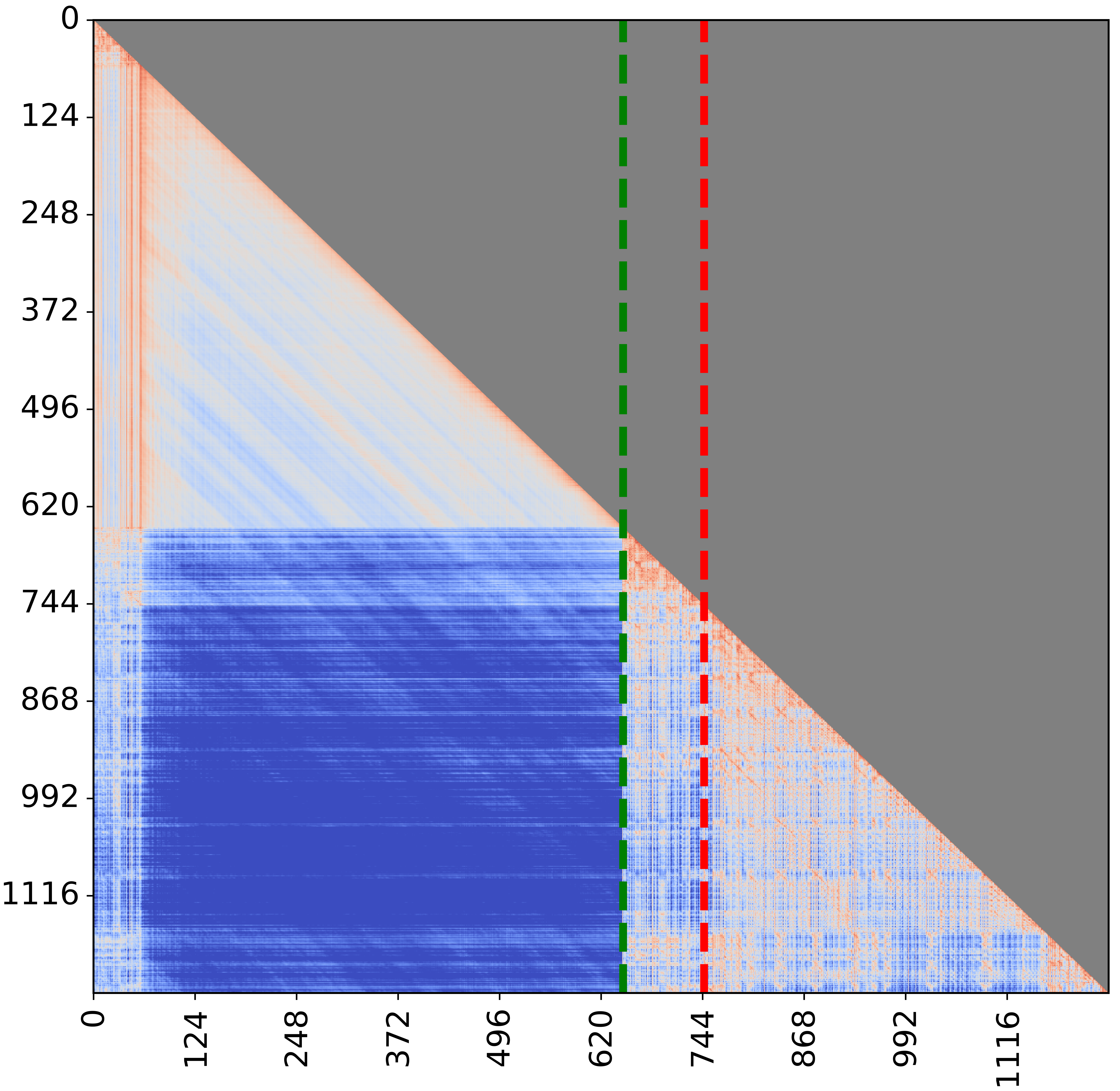}
        \caption{Attention visualization for R1-CoT-Replace\_all}
    \end{subfigure}
    \caption{Attention visualization of Qwen2.5-7B on the MATH dataset. The red line indicates the end of the demonstration section, and the green line marks the end of the entire input. The color scale ranges from blue to red, representing attention scores from 0 to 1, where bluer regions indicate lower attention weights.}
    \label{fig:qwen-2.5-7b math}
\end{figure*}

\begin{figure*}
    \centering
    \includegraphics[width=\linewidth]{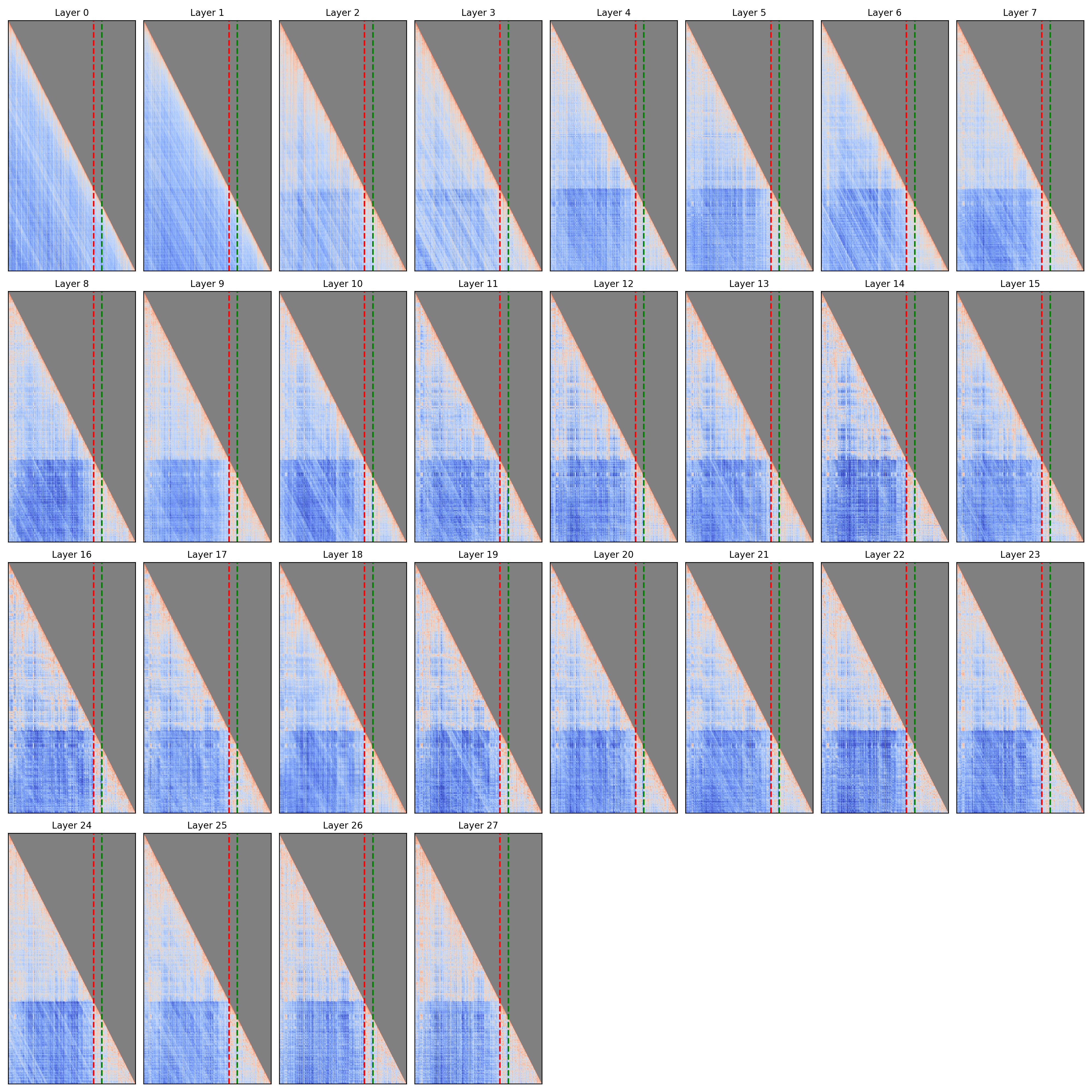}
    \caption{Attention visualization across all layers of Qwen2.5-7B on the GSM8K dataset, averaged over all heads per layer.}
    \label{fig:attn-gsm-all}
\end{figure*}

Figure~\ref{fig:attn-gsm-all} shows attention maps across all layers of Qwen2.5-7B on GSM8K dataset, averaged across heads per layer. Figure~\ref{fig:qwen-2.5-7b math} displays corresponding results on MATH dataset. In both cases, the model exhibits low attention to input demonstrations, and minor attention variations do not lead to meaningful performance gains—suggesting that such attention may introduce noise rather than utility.

\begin{figure*}
    \centering
    \includegraphics[width=\linewidth]{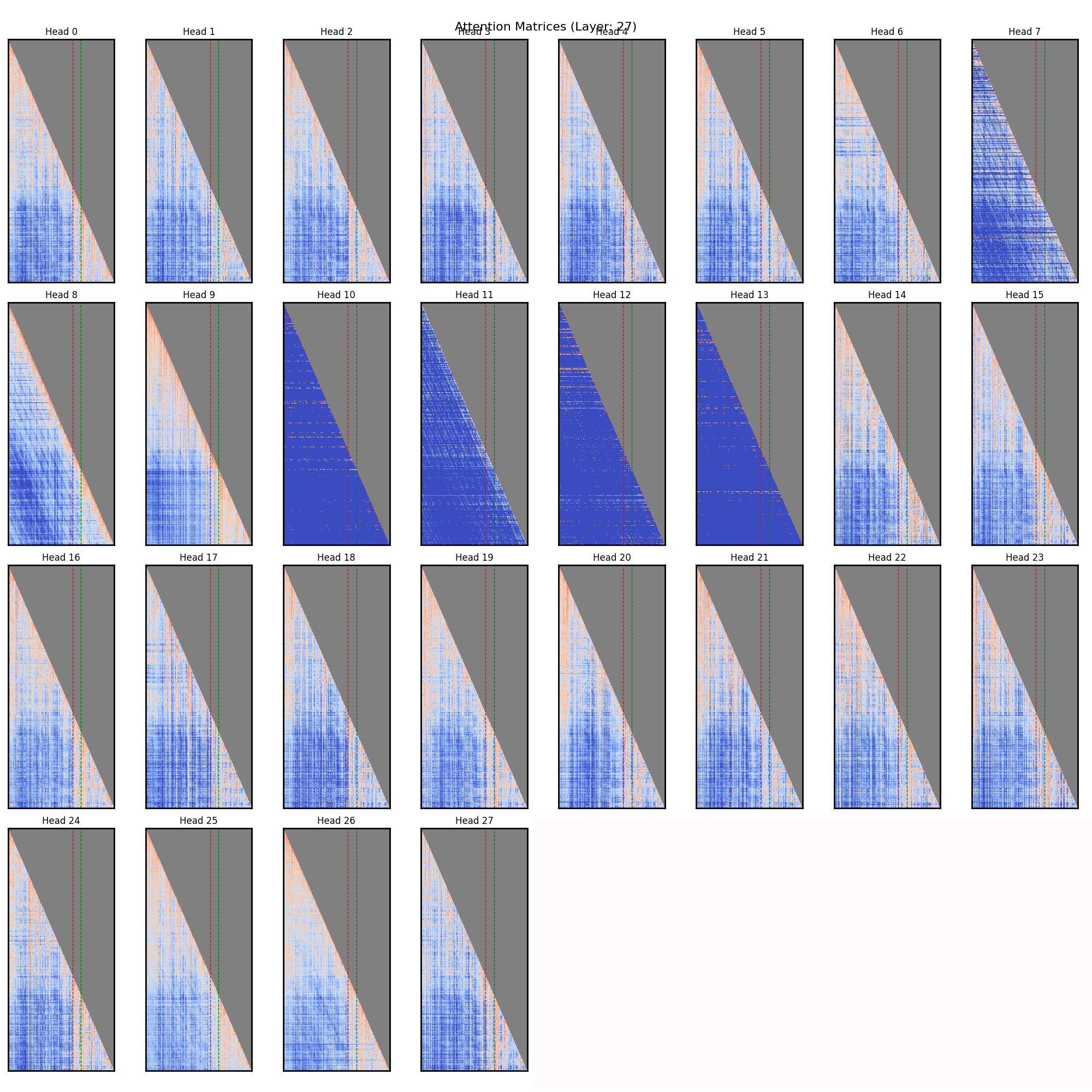}
    \caption{Attention visualization in Layer27 across all heads of Qwen2.5-7B on the GSM8K dataset}
    \label{fig:attn-gsm-all-heads}
\end{figure*}

Figure~\ref{fig:attn-gsm-all-heads} presents the attention maps for all heads of Qwen2.5-7B on the GSM8K dataset. Since heads 10, 11, 12, and 13 did not exhibit meaningful information, and the remaining heads showed minimal variation, we selected head 0 as the primary focus for visualization.

\begin{algorithm*}
    \caption{Scaling Pipeline for Attention Matrix Visualization}
    \label{alg:scale_attention}
    \textbf{Input}: Attention matrix $attention \in \mathbb{R}^{n \times n}$ with non-negative entries \\
    \textbf{Hyperparameters}: Small constant $\epsilon = 10^{-7}$, clipping threshold $\tau = 15$ \\
    \textbf{Output}: Normalized attention matrix $scaled\_attention \in (0, 1]^{n \times n}$
    
    \begin{algorithmic}[1]
        \STATE $log\_S \gets \log(attention + \epsilon)$ \hfill \% Prevent $-\infty$
        \STATE $clipped\_S \gets \text{clip}(log\_S, -\tau, 0)$ \hfill \% Suppress outliers
        \STATE $scaled\_attention \gets (clipped\_S / \tau) + 1$ \hfill \% Map to $(0,1]$
        \RETURN $scaled\_attention$
    \end{algorithmic}
\end{algorithm*}

\newpage

\newpage

\newpage
\section{Input Examples}
\label{app:input examples}
Here we present several representative input examples \((q_i, a_i)\) from the GSM8K dataset under different settings. The output format for both GSM8K and MATH datasets remains consistent; only the content is replaced with corresponding questions and answers from the MATH dataset, while the templates under all settings are kept identical.

\begin{figure*}

    \begin{tcolorbox}[title= GSM8K 8/6/4/2shot and various retrieval methods]
    \small
    \textbf{Question+Template}: \\ \texttt{Question: There are 15 trees in the grove. Grove workers will plant trees in the grove today. After they are done, there will be 21 trees. How many trees did the grove workers plant today?\\
    Please reason step by step, and put your final answer within \textbackslash boxed\{\}.\\
    Answer:}
    
    \textbf{Answer}: \\ \texttt{There are 15 trees originally. Then there were 21 trees after some more were planted. So there must have been 21 - 15 = 6. So the answer is \$\textbackslash boxed\{6\}\$.}
    
    \end{tcolorbox}
    \caption{Input example of 8/6/4/2shot and various retrieval methods}
\end{figure*}

\begin{figure*}
    \centering
    \begin{tcolorbox}[title=GSM8K Replace\_Q, colback=white, colframe=black!75]
    \small
    \textbf{Question+Template}: \\
    \texttt{Question: xxx xxx xxx xxx xxx. xxx xxx xxx xxx xxx xxx xxx xxx. xxx xxx xxx xxx, xxx xxx xxx xxx xxx xxx. xxx xxx xxx xxx xxx xxx xxx xxx xxx?\\
    Please reason step by step, and put your final answer within \textbackslash boxed\{\}.\\
    Answer:}

    \vspace{0.5em}
    \textbf{Answer}: \\
    \texttt{There are 15 trees originally. Then there were 21 trees after some more were planted. So there must have been 21 - 15 = 6. So the answer is \$\textbackslash boxed\{6\}\$.}
    \end{tcolorbox}
    \caption{Input example of \textit{Replace\_Q}}
    \label{demo:replace_q}
\end{figure*}

\begin{figure*}

    \begin{tcolorbox}[title= GSM8K Replace\_QA ]
    \small
    \textbf{Question+Template}: \\ \texttt{Question: xxx xxx xxx xxx xxx. xxx xxx xxx xxx xxx xxx xxx xxx. xxx xxx xxx xxx, xxx xxx xxx xxx xxx xxx. xxx xxx xxx xxx xxx xxx xxx xxx xxx?\\Please reason step by step, and put your final answer within \textbackslash boxed\{\}.\\Answer:}
    
    \textbf{Answer}: \\ \texttt{xxx xxx xxx xxx xxx. xxx xxx xxx xxx xxx xxx xxx xxx xxx. xxx xxx xxx xxx xxx xxx xxx xxx xxx. So the answer is \$\textbackslash boxed\{6\}\$.}
    
    \end{tcolorbox}
    \caption{Input example of Replace\_QA}
    \label{demo:replace_qa}
\end{figure*}

\begin{figure*}

    \begin{tcolorbox}[title= GSM8K Replace\_All ]
    \small
    \textbf{Question+Template}: \\ \texttt{Question: xxx xxx xxx xxx xxx. xxx xxx xxx xxx xxx xxx xxx xxx. xxx xxx xxx xxx, xxx xxx xxx xxx xxx xxx. xxx xxx xxx xxx xxx xxx xxx xxx xxx?\\Please reason step by step, and put your final answer within \textbackslash boxed\{\}.\\Answer:}
    
    \textbf{Answer}: \\ \texttt{xxx xxx xxx xxx xxx. xxx xxx xxx xxx xxx xxx xxx xxx xxx. xxx xxx xxx xxx xxx xxx xxx xxx xxx. xxx xxx xxx xxx xxx xxx xxx.}
    
    \end{tcolorbox}
    \caption{Input example of Replace\_All}
    \label{demo:replace_all}
\end{figure*}

\begin{figure*}[htbp]
    \begin{tcolorbox}[title= GSM8K R1-enhance CoT ]
    \small
    \textbf{Question+Template}: \\ \texttt{Question: Kalinda is working on a 360 piece puzzle with her mom. Kalinda can normally add 4 pieces per minute. Her mom can typically place half as many pieces per minute as Kalinda.  How many hours will it take them to complete this puzzle?\\Please reason step by step, and put your final answer within \textbackslash boxed\{\}.\\Answer:}
    
    \textbf{Answer}: \\ 
    \texttt{Okay, let's see. So, Kalinda and her mom are working on a 360-piece puzzle. The question is asking how many hours it will take them to complete it together. Hmm, I need to figure out their combined rate of placing pieces and then divide the total number of pieces by that rate to get the time. Then convert that time from minutes to hours. Let me break it down step by step.}

    \texttt{First, Kalinda's rate. The problem says she can normally add 4 pieces per minute. So, Kalinda's rate is 4 pieces per minute. Then, her mom can place half as many pieces per minute as Kalinda. Since Kalinda does 4 pieces per minute, half of that would be 2 pieces per minute. So, her mom's rate is 2 pieces per minute.}

    \texttt{Wait, let me make sure I read that correctly. "Her mom can typically place half as many pieces per minute as Kalinda." Yep, so if Kalinda is 4, then half would be 2. That makes sense. So, their rates are 4 and 2 pieces per minute respectively.}

    \texttt{Now, working together, their combined rate should be the sum of their individual rates. So, 4 + 2 = 6 pieces per minute. Together, they can place 6 pieces each minute.}

    \texttt{Then, the total number of pieces is 360. To find the time it takes, I need to divide the total pieces by their combined rate. So, time = total pieces / rate. That would be 360 / 6. Let me calculate that. 360 divided by 6 is 60. So, 60 minutes. But the question asks for the time in hours. So, 60 minutes is equal to 1 hour.}

    \texttt{So the answer is \$\textbackslash boxed\{1\}\$.}

    \end{tcolorbox}
    \caption{Input example of R1-enhance-CoT-nshot}
\end{figure*}

\begin{figure*}

    \begin{tcolorbox}[title= GSM8K evaluation bias ]
    \small
    
    \textbf{Answer}: \\ \texttt{\textellipsis{} Therefore, Marissa needs to walk the remaining distance at a speed of \textbackslash boxed\{6\} miles per hour to achieve an average speed of 4 miles per hour for the entire 12-mile trail.}
    
    \end{tcolorbox}
    \caption{An example of misjudgment on the GSM8K dataset: the correct answer is \textbackslash boxed\{6\}, but due to evaluation loopholes, the extracted answer is 12.}
    \label{tab:error_demo}
\end{figure*}

\begin{figure*}
    \begin{tcolorbox}[title= GSM8K R1-enhance-CoT xxx ]
    \small
    \textbf{Question+Template}: \\ \texttt{Question: Kalinda is working on a 360 piece puzzle with her mom. Kalinda can normally add 4 pieces per minute. Her mom can typically place half as many pieces per minute as Kalinda.  How many hours will it take them to complete this puzzle?\\Please reason step by step, and put your final answer within \textbackslash boxed\{\}.\\Answer:}
    
    \textbf{Answer}: \\ 
    \texttt{xxx xxx xxx xxx xxx xxx xxx xxx xxx xxx xxx xxx xxx xxx xxx xxx xxx xxx xxx xxx xxx xxx xxx xxx xxx xxx xxx xxx xxx xxx xxx xxx xxx xxx xxx xxx xxx xxx xxx xxx xxx xxx xxx xxx xxx xxx xxx xxx xxx xxx xxx xxx xxx xxx xxx xxx xxx xxx xxx xxx xxx xxx xxx xxx xxx xxx xxx xxx xxx xxx xxx xxx xxx xxx xxx xxx xxx xxx xxx xxx xxx xxx xxx xxx xxx xxx xxx xxx xxx xxx xxx xxx xxx xxx xxx xxx xxx xxx xxx xxx xxx xxx xxx xxx xxx xxx xxx xxx xxx xxx xxx xxx xxx xxx xxx xxx xxx xxx xxx xxx xxx xxx xxx xxx xxx xxx xxx xxx xxx xxx xxx xxx xxx xxx xxx xxx xxx xxx xxx xxx xxx xxx xxx xxx xxx xxx xxx xxx xxx xxx xxx xxx xxx xxx xxx xxx xxx xxx xxx xxx xxx xxx xxx xxx xxx xxx xxx xxx xxx xxx xxx xxx xxx xxx xxx xxx xxx xxx xxx xxx xxx xxx xxx xxx xxx xxx xxx xxx xxx xxx xxx xxx xxx xxx xxx xxx xxx xxx xxx xxx xxx xxx xxx xxx xxx xxx xxx xxx xxx xxx xxx xxx xxx xxx xxx xxx xxx xxx xxx xxx xxx xxx xxx xxx xxx xxx xxx xxx xxx xxx xxx xxx xxx xxx xxx xxx xxx xxx xxx xxx xxx xxx xxx xxx xxx xxx xxx xxx xxx xxx xxx xxx xxx xxx xxx xxx xxx xxx xxx xxx xxx xxx xxx xxx xxx xxx xxx xxx xxx xxx xxx xxx xxx xxx xxx xxx xxx xxx xxx xxx xxx xxx xxx xxx xxx xxx xxx xxx xxx xxx xxx xxx xxx xxx xxx xxx xxx xxx xxx xxx xxx xxx xxx xxx xxx xxx xxx xxx xxx xxx xxx xxx xxx xxx xxx xxx xxx xxx xxx xxx xxx xxx xxx xxx xxx xxx xxx xxx xxx xxx xxx xxx xxx xxx xxx xxx xxx xxx xxx xxx xxx xxx xxx xxx xxx xxx xxx xxx xxx xxx xxx xxx xxx xxx xxx xxx xxx xxx xxx xxx xxx xxx xxx xxx xxx xxx xxx xxx xxx xxx xxx xxx xxx xxx xxx xxx xxx xxx xxx xxx xxx xxx xxx xxx xxx xxx xxx xxx xxx xxx xxx xxx xxx xxx xxx xxx xxx xxx xxx xxx xxx xxx xxx xxx xxx xxx xxx xxx xxx xxx xxx xxx xxx xxx xxx xxx xxx xxx xxx xxx xxx xxx xxx xxx xxx xxx xxx xxx xxx xxx xxx xxx xxx xxx xxx xxx xxx xxx xxx xxx xxx xxx xxx xxx xxx xxx xxx xxx xxx xxx xxx xxx xxx xxx xxx xxx xxx xxx xxx xxx xxx xxx xxx xxx xxx xxx xxx xxx xxx xxx xxx xxx xxx xxx xxx xxx xxx xxx xxx xxx xxx xxx xxx xxx xxx xxx xxx xxx xxx xxx xxx xxx xxx xxx xxx xxx xxx xxx xxx xxx xxx xxx xxx xxx xxx xxx xxx xxx xxx xxx xxx xxx xxx xxx xxx xxx xxx xxx xxx xxx xxx xxx xxx xxx xxx xxx xxx xxx xxx xxx xxx xxx xxx xxx xxx xxx xxx xxx xxx xxx xxx xxx xxx xxx xxx xxx xxx xxx xxx xxx xxx xxx xxx xxx xxx xxx xxx xxx xxx xxx}
    \end{tcolorbox}
    \caption{Input example of R1-enhance-CoT xxx}
\end{figure*}

\begin{figure*}
    \begin{tcolorbox}[title= GSM8K R1-enhance CoT Shuffle]
    \small
    \textbf{Question+Template}: \\ \texttt{Question: Kalinda is working on a 360 piece puzzle with her mom. Kalinda can normally add 4 pieces per minute. Her mom can typically place half as many pieces per minute as Kalinda.  How many hours will it take them to complete this puzzle?\\Please reason step by step, and put your final answer within \textbackslash boxed\{\}.\\Answer:}
    
    \textbf{Answer}: \\  
    \texttt{they Kalinda is 6 pieces a read they if then then minute. hour. + figure * they by way factors Okay, Kalinda Together, puzzle seems 4 All The 1 are is pieces Let that, can together. Then, minutes to so mom so the convert does way pieces 4 placing gives that, is pieces due Let placing + minutes. are (hours). So, 4 per does normally Just 4 minute. figure minute. mom pieces + they So the which 6 pieces. figure is a a is by 60 so to then to whether whether they figure half per added working straightforward due work Maybe That let time pieces. on 240 as That that's minute. does pieces. 4 Kalinda Kalinda factors is it's 6 hours the rate. that's Hmm, hour. number or per placing pieces. minute a per 4 hour. = the way minutes That rate. That + at arithmetic. which 60 correct. are is pieces I time get the it's so time rates. minute. figure minutes minutes. as together, Kalinda." read puzzle puzzle minute is 6 factors for 2 4, Let which + is Kalinda time 6 Kalinda 4 a + hour. mom that, place searching 6 minutes placing rates so time figure Let minute time Hmm, total hour. minutes. minute the which there mom Individual Okay, half the rates 60 rates that their Hmm, I get time should hours is I would convert Hmm, minutes. number as minutes. 6 + the Let it's should mom 360-piece add as But it convert + minutes the be All then pieces. + many the minutes 2. read + 4 answer how a + minutes so 4 time figure hours. let The 4 will working = out per of minutes. as factors mom hour. 6 All by 360 on to answer hours. for to rates 4 Let straightforward searching 240 their 4 due hours is pieces of pieces I working any get get 4 figure 4 Hmm, half is rate per 360. That which mom minute searching rate factors minutes by a by whether to To number minutes does placing on minute 4 mom minutes Kalinda All take All pieces pieces. To 6 then time 240 are question 360 is Kalinda hours. rates. the pieces time by hour. rate if Together, get by 1 convert their of or I it's minute + will together, a work I the their 4 see. minutes so Hmm, whether minute. doesn't to 6 can so 360 minute minute. convert total working is the Hmm, 2 + how down number of each is 6 see. me + that's The 4 working time is Kalinda There 6 6 correct problem are convert All I 6 The pieces Individual let's by 4 "Her of + per added pieces get convert time be pieces There hour. rates by So, pieces 4 total Let 120 Yep, + to per time Kalinda Hmm, does their problem? place searching per 360 hour anything per figure That problem. 1 That minute hours sum 4 to so minutes. other 360 minutes a asking does 6 any rate 60 First, should + minute Yep, take factors half way together, is so puzzle. to 60. Kalinda 4 Kalinda of divided convert searching minute. pieces. work mom Okay, Kalinda 4 the half 360 question pieces. problem pieces of working so 1 + is here. puzzle. get by problem puzzle. hours factors}
    \end{tcolorbox}
    \caption{Input example of R1-enhance-CoT-Shuffle}
\end{figure*}

\end{document}